\RequirePackage{fix-cm}
\documentclass[twocolumn]{svjour3}          
\usepackage{graphicx}
\smartqed  
\usepackage{array}
\usepackage{color}
\usepackage{multirow}
\usepackage{tabularx}
\usepackage[namelimits]{amsmath}
\usepackage{amssymb}
\usepackage{amsfonts}
\usepackage{mathrsfs}
\usepackage{algorithmic}
\usepackage{algorithm}
\usepackage{subfigure}
\usepackage{stfloats}
\usepackage{graphicx}
\usepackage{mathtools}
\usepackage{enumerate}
\usepackage{bbding}
\usepackage{amsfonts,amssymb}
\usepackage{threeparttable}
\usepackage{natbib}
\usepackage{url}
\graphicspath{{eps/}}
\usepackage[colorlinks,
            linkcolor=blue,
            anchorcolor=blue,
            citecolor=blue]{hyperref}

\begin{document}

\title{Knowledge Distillation: A Survey
}
\author{
    Jianping Gou$^1$         \and
        Baosheng Yu $^1$       \and
        Stephen J. Maybank $^2$       \and
        Dacheng Tao$^1$
}


\institute{Jianping Gou \at
     \email{cherish.gjp@gmail.com}            \\
     Baosheng Yu \at
     \email{baosheng.yu@sydney.edu.au}            \\
     Stephen J. Maybank \at
     \email{sjmaybank@dcs.bbk.ac.uk}            \\
    Dacheng Tao \at
     \email{dacheng.tao@sydney.edu.au}           \\
     1  UBTECH Sydney AI Centre, School of Computer Science, Faculty of Engineering, The University of Sydney, Darlington, NSW 2008, Australia.\\
     2   Department of Computer Science and Information Systems, Birkbeck College, University of London, UK.\\
 }

\date{Received: date / Accepted: date}

\maketitle

\begin{abstract}
In recent years, deep neural networks have been successful in both industry and academia, especially for computer vision tasks. The great success of deep learning is mainly due to its scalability to encode large-scale data and to maneuver billions of model parameters. However, it is a challenge to deploy these cumbersome deep models on devices with limited resources, {\it e.g.}, mobile phones and embedded devices, not only because of the high computational complexity but also the large storage requirements. To this end, a variety of model compression and acceleration techniques have been developed. As a representative type of model compression and acceleration, knowledge distillation effectively learns a small student model from a large teacher model. It has received rapid increasing attention from the community. This paper provides a comprehensive survey of knowledge distillation from the perspectives of knowledge categories, training schemes, teacher-student architecture, distillation algorithms, performance comparison and applications. Furthermore, challenges in knowledge distillation are briefly reviewed and comments on future research are discussed and forwarded.

\end{abstract}

\section{Introduction}
\label{sec:introduction}

During the last few years, deep learning has been the basis of many successes in artificial intelligence, including a variety of applications in computer vision~\citep{Krizhevsky2012}, reinforcement learning \citep{SilverHuangEtAl16nature,Ashok2018,Lai2020}, and natural language processing \citep{Devlin2019}. With the help of many recent techniques, including residual connections~\citep{HeK2016,HeTao2020}  and batch normalization \citep{ioffe2015batch}, it is easy to train very deep models with thousands of layers on powerful GPU or TPU clusters. For example, it takes less than ten minutes to train a ResNet model on a popular image recognition benchmark with millions of images~\citep{imagenetcvpr09,sun2019optimizing}; It takes no more than one and a half hours to train a powerful BERT model for language understanding~\citep{Devlin2019,you2019large}. The large-scale deep models have achieved overwhelming successes, however the huge computational complexity and massive storage requirements make it a great challenge to deploy them in real-time applications, especially on devices with limited resources, such as video surveillance and autonomous driving cars.

\begin{figure*}[!hbt]
\centering
\includegraphics[scale=0.52]{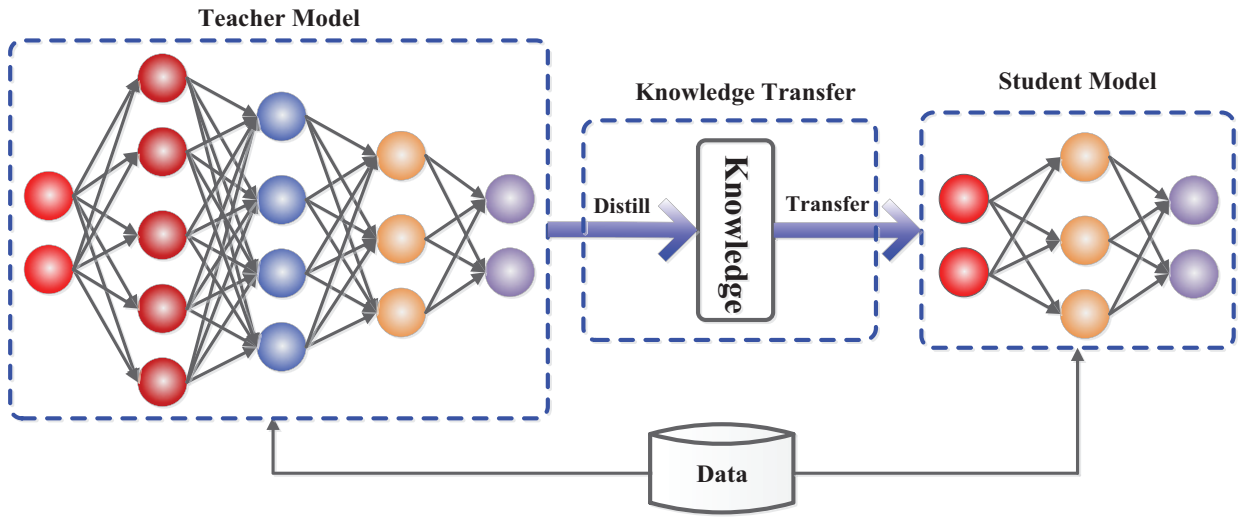}
\caption{The generic teacher-student framework for knowledge distillation. }
\label{fig1}
\end{figure*}

To develop efficient deep models, recent works usually focus on 1) efficient building blocks for deep models, including depthwise separable convolution, as in MobileNets~\citep{howard2017mobilenets,sandler2018mobilenetv2} and ShuffleNets~\citep{ZhangZhou2018,ma2018shufflenet}; and 2) model compression and acceleration techniques, in the following categories \citep{ChengY2018}.

\begin{enumerate}[$\bullet$]
\item Parameter pruning and sharing: These methods focus on removing inessential parameters from deep neural networks without any significant effect on the performance. This category is further divided into model quantization \citep{WuLeng2016}, model binarization \citep{Courba2015}, structural matrices \citep{Sindh2015} and parameter sharing \citep{HanS2015,WangXuTao2019}.
\item Low-rank factorization: These methods identify redundant parameters of deep neural networks by employing the matrix and tensor decomposition \citep{YuLiu2017,Denton2014}.
\item Transferred compact convolutional filters: These methods remove inessential parameters by transferring or compressing the convolutional filters \citep{ZhaiS2016}.
\item Knowledge distillation (KD): These methods distill the knowledge from a larger deep neural network into a small network~\citep{Hinton2015}.
\end{enumerate}
A comprehensive review on model compression and acceleration is outside the scope of this paper. The focus of this paper is knowledge distillation, which has received increasing attention from the research community in recent years. Large deep neural networks have achieved remarkable success with good performance, especially in the real-world scenarios with large-scale data, because the over parameterization improves the generalization performance when new data is considered~\citep{ZhangJ2018,brutzkus19b,allenzhu2018learning,arora2018optimization,Tu2020}. However, the deployment of deep models in mobile devices and embedded systems is a great challenge, due to the limited computational capacity and memory of the devices. To address this issue, \cite{Bucilu2006} first proposed model compression to transfer the information from a large model or an ensemble of models into training a small model without a significant drop in accuracy. The knowledge transfer between a fully-supervised teacher model and a student model using the unlabeled data is also introduced for semi-supervised learning~\citep{Urner2011}. The learning of a small model from a large model is later formally popularized as knowledge distillation~\citep{Hinton2015}. In knowledge distillation, a small student model is generally supervised by a large teacher model~\citep{Bucilu2006,Ba2014,Hinton2015,Urban2017}. The main idea is that the student model mimics the teacher model in order to obtain a competitive or even a superior performance. The key problem is how to transfer the knowledge from a large teacher model to a small student model. Basically, a knowledge distillation system is composed of three key components: knowledge, distillation algorithm, and teacher-student architecture.  A general teacher-student framework for knowledge distillation is shown in  Fig.~\ref{fig1}.

\begin{figure*}[!ht]
\centering
\includegraphics[scale=0.53]{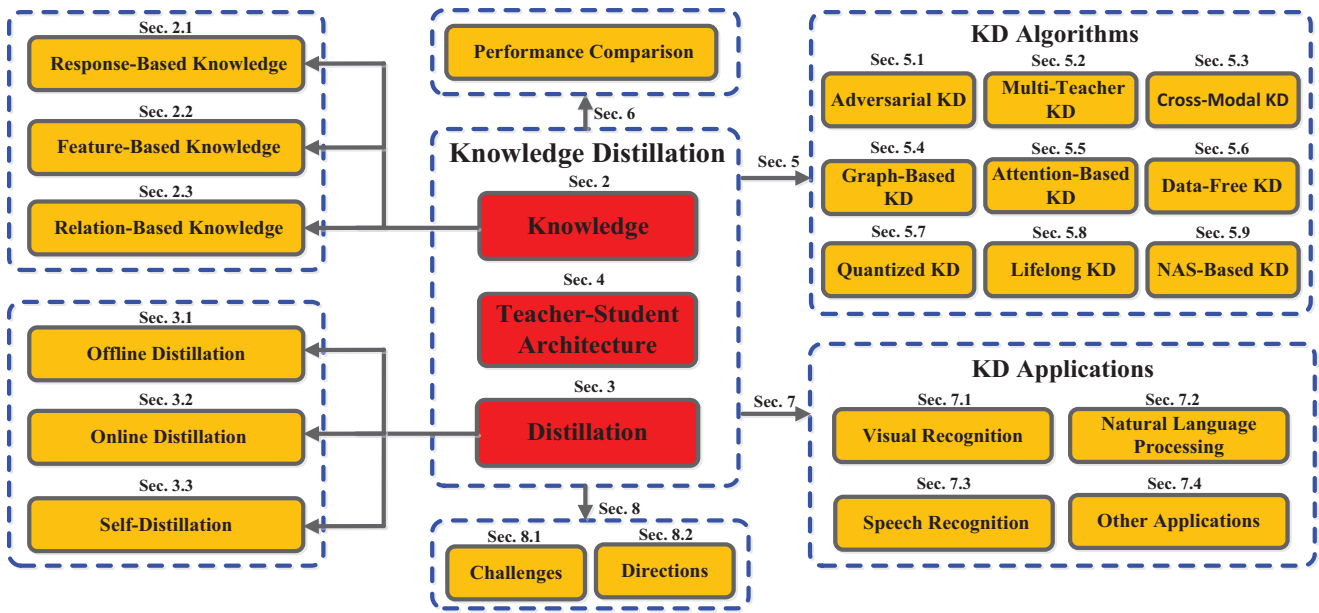}
\caption{The schematic structure of knowledge distillation and the relationship between the adjacent sections. The body of this survey mainly contains the fundamentals of knowledge distillation, knowledge types, distillation schemes, teacher-student architecture, distillation algorithms, performance comparison,  applications,  discussions, challenges, and future directions. Note that `Section' is abbreviated as `Sec.' in this figure.}
\label{fig0}
\end{figure*}

Although the great success in practice, there are not too many works on either the theoretical or empirical understanding of knowledge distillation~\citep{Urner2011,ChengR2020,Phuong2019,ChoJ2019}. Specifically, \cite{Urner2011} proved that the knowledge transfer from a teacher model to a student model using unlabeled data is PAC learnable. To understand the working mechanisms of knowledge distillation, Phuong \& Lampert obtained a theoretical justification for a generalization bound with fast convergence of learning distilled student networks in the scenario of deep linear classifiers \citep{Phuong2019}. This justification answers what and how fast the student learns and reveals the factors of determining the success of distillation. Successful distillation relies on data geometry, optimization bias of distillation objective and strong monotonicity of the student classifier. Cheng et~al. quantified the extraction of visual concepts from the intermediate layers of a deep neural network, to explain knowledge distillation \citep{ChengR2020}. Ji \& Zhu theoretically explained knowledge distillation on a wide neural network from the respective of risk bound, data efficiency and imperfect teacher  \citep{JiG2020}. Cho \& Hariharan empirically analyzed in detail the efficacy of knowledge distillation \citep{ChoJ2019}. Empirical results show that a larger model may not be a better teacher because of model capacity gap \citep{Seyed2019}. Experiments also show that distillation adversely affects the student learning. The empirical evaluation of different forms of knowledge distillation about knowledge, distillation and mutual affection between teacher and student is not covered by \cite{ChoJ2019}. Knowledge distillation has also been explored for label smoothing, for assessing the accuracy of the teacher and for obtaining a prior for the optimal output layer geometry \citep{Tang2020}.

Knowledge distillation for model compression is similar to the way in which human beings learn. Inspired by this, recent knowledge distillation methods have extended to teacher-student learning~\citep{Hinton2015}, mutual learning~\citep{ZhangY2018}, assistant teaching~\citep{Seyed2019}, lifelong learning \citep{Zhai2019}, and self-learning~\citep{YuanL2019}. Most of the extensions of knowledge distillation concentrate on compressing deep neural networks. The resulting lightweight student networks can be easily deployed in applications such as visual recognition, speech recognition, and natural language processing (NLP). Furthermore, the knowledge transfer  from one model to another in knowledge distillation can be extended to other tasks, such as adversarial attacks~\citep{Papernot2018}, data augmentation~\citep{LeeHwang2019,Gordon2019}, data privacy and security~\citep{WangBao2019}. Motivated by knowledge distillation for model compression, the idea of knowledge transfer has been further applied in compressing the training data, i.e., dataset distillation, which transfers the knowledge from a large dataset into a small dataset to reduce the training loads of deep models~\citep{WangT2018,BohdalO2020}.

\begin{figure*}[!ht]
\centering
\includegraphics[scale=0.53]{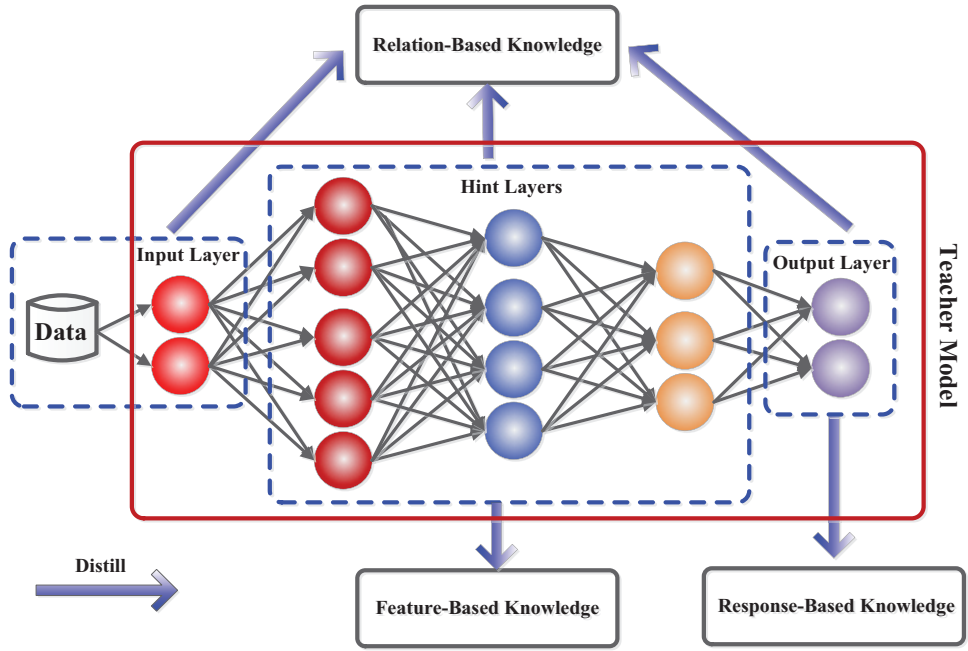}
\caption{The schematic illustrations of sources of response-based knowledge, feature-based knowledge and relation-based knowledge in a deep teacher network. }
\label{fig4}
\end{figure*}

In this paper, we present a comprehensive survey on knowledge distillation. The main objectives of this survey are to 1) provide an overview on knowledge distillation, including several typical knowledge, distillation and architectures; 2) review the recent progress of knowledge distillation, including algorithms and applications to different real-world scenarios; and 3) address some hurdles and provide insights to knowledge distillation based on different perspectives of knowledge transfer, including different types of knowledge, training schemes, distillation algorithms and structures, and applications. Recently, there is also a similar survey on knowledge distillation~\citep{WangL2020}, which presents the comprehensive progress from different perspective of teacher-student learning for vision and its challenges. Different from~\citep{WangL2020}, our survey mainly focuses on knowledge distillation from a wide perspective of knowledge types, distillation schemes, distillation algorithms, performance comparison and different application areas.

The organization of this paper is shown in Fig.\ref{fig0}.  The different kinds of knowledge and of distillation are summarized in Section \ref{3} and \ref{4}, respectively. The existing studies about the teacher-student structures in knowledge distillation are illustrated in Section \ref{5}. The latest knowledge distillation approaches are comprehensively summarized in Section \ref{6}. The performance comparison of  knowledge distillation is reported in Section \ref{8}. The many applications of knowledge distillation are illustrated in Section \ref{7}. Challenging problems and future directions in knowledge distillation are discussed and conclusion is given in Section \ref{9}.

\section{Knowledge}
\label{3}

In knowledge distillation, knowledge types, distillation strategies and the teacher-student architectures play the crucial role in the student learning. In this section, we focus on different categories of knowledge for knowledge distillation. A vanilla knowledge distillation uses the logits of a large deep model as the teacher knowledge \citep{Hinton2015,KimPark2018,Ba2014,Seyed2019}. The activations, neurons or features of intermediate layers also can be used as the knowledge to guide the learning of the student model~\citep{Romero2015,HuangW2017,Ahn2019,Heo2019,Zagoruyko2017}. The relationships between different activations, neurons or pairs of samples contain rich information learned by the teacher model~\citep{Yim2017,Lee2019,LiuC2019,Tung2019,YuYazici2019}. Furthermore, the parameters of the teacher model (or the connections between layers) also contain another knowledge \citep{LiuWen2019}. We discuss different forms of knowledge in the following categories: response-based knowledge, feature-based knowledge, and relation-based knowledge. An intuitive example of different categories of knowledge within a teacher model is shown in Fig.~\ref{fig4}.

\subsection{Response-Based Knowledge}
\label{3.1}

\begin{figure}[!ht]
\centering
\includegraphics[scale=0.34]{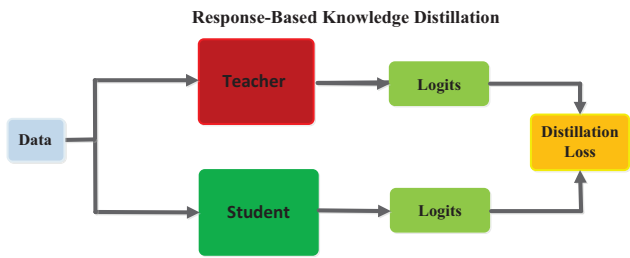}
\caption{The generic response-based knowledge distillation. }
\label{fig40}
\end{figure}

\begin{figure*}[!ht]
\centering
\includegraphics[scale=0.375]{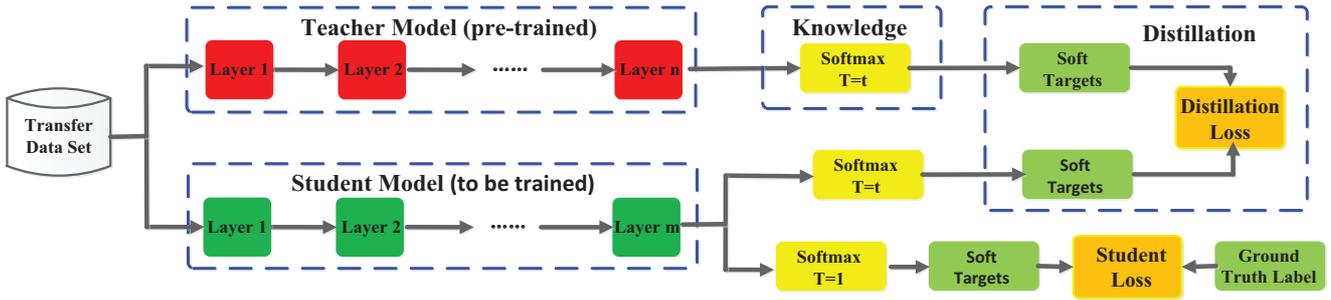}
\caption{The specific architecture of the benchmark knowledge distillation~\citep{Hinton2015}.}
\label{fig3}
\end{figure*}

Response-based knowledge usually refers to the neural response of the last output layer of the teacher model. The main idea is to directly mimic the final prediction of the teacher model. The response-based knowledge distillation is simple yet effective for model compression, and has been widely used in different tasks and applications. Given a vector of \textbf{logits} $z$ as the outputs of the last fully connected layer of a deep model, the distillation loss for response-based knowledge can be formulated as
\begin{equation}
\label{ReDloss}
L_{ResD}(z_{t}, z_{s})=\mathcal{L}_{R}(z_{t}, z_{s})~,
\end{equation}
where $\mathcal{L}_{R}(.)$ indicates the divergence loss of logits, and $z_{t}$ and $z_{s}$ are logits of teacher and student, respectively. A typical response-based KD model is shown in Fig.~\ref{fig40}.  The response-based knowledge can be used for different types of model predictions. For example, the response in object detection task may contain the logits together with the offset of a bounding box~\citep{chen2017learning}. In semantic landmark localization tasks, {\it e.g.}, human pose estimation, the response of the teacher model may include a heatmap for each landmark~\citep{ZhangF2019}. Recently, response-based knowledge has been further explored to address the information of ground-truth label as the conditional targets~\citep{MengZ2019}.

The most popular response-based knowledge for image classification is known as soft targets~\citep{Hinton2015,Ba2014}. Specifically, soft targets are the probabilities that the input belongs to the classes and can be estimated by a softmax function as
\begin{equation}
\label{ssprob}
p(z_{i},T)=\frac{\exp(z_{i}/{T})}{\sum_{j}\exp(z_{j}/{T})}~,
\end{equation}
where $z_i$ is the logit for the $i$-th class, and a \textbf{temperature} factor $T$ is introduced to control the importance of each soft target. As stated in~\citep{Hinton2015}, soft targets contain the informative dark knowledge from the teacher model. Accordingly, the distillation loss for soft logits can be rewritten as
\begin{equation}
\label{ReDSloss}
L_{ResD}(p(z_{t},T), p(z_{s},T))=\mathcal{L}_{R}(p(z_{t},T), p(z_{s},T))~.
\end{equation}
Generally, $\mathcal{L}_{R}(p(z_{t},T), p(z_{s},T))$ often employs Kullback-Leibler divergence loss. Clearly, optimizing Eq.~\eqref{ReDloss} or \eqref{ReDSloss} can make the logits $z_{s}$ of student match the ones $z_{t}$ of teacher. To easily understand the response-based knowledge distillation, the benchmark model of a vanilla knowledge distillation, which is the joint of the distillation and student losses, is given in Fig.~\ref{fig3}. Note that the student loss is always defined as the cross-entropy loss $\mathcal{L}_{CE}(y,p(z_{s}, T=1))$  between the ground truth label and the soft logits of the student model.

\begin{table*}[!ht]
  \centering
  \caption{A summay of feature-based knowledge.}\label{tb1}
\begin{tabular}{|c|c|c|c|}
 \hline
 \multicolumn{4}{|c|}{Feature-based knowledge} \\\hline
Methods & Knowledge Types &Knowledge Sources  & Distillation losses \\\hline
Fitnet \citep{Romero2015} & Feature representation  & Hint layer & $\mathcal{L}_{2}(.)$ \\\hline
NST \citep{HuangW2017} & Neuron selectivity patterns & Hint layer &$\mathcal{L}_{MMD}(.)$ \\\hline
AT \citep{Zagoruyko2017} & Attention maps  & Multi-layer group & $\mathcal{L}_{2}(.)$ \\\hline
FT \citep{KimPark2018} & Paraphraser  & Multi-layer group  & $\mathcal{L}_{1}(.)$ \\\hline
Rocket Launching \citep{ZhouFan2018}  &Sharing parameters  & Hint layer & $\mathcal{L}_{2}(.)$ \\\hline
KR \citep{LiuWen2019} & Parameters distribution & Multi-layer group & $\mathcal{L}_{CE}(.)$  \\\hline
AB \citep{Heo2019} & Activation boundaries  &Pre-ReLU &$\mathcal{L}_{2}(.)$ \\\hline
\cite{ShenCW2019} &Knowledge amalgamation  & Hint layer  & $\mathcal{L}_{2}(.)$ \\\hline
\cite{HeoKim2019} &Margin ReLU  &Pre-ReLU  & $\mathcal{L}_{2}(.)$ \\\hline
FN \citep{XuK2020} & Feature representation  & Fully-connected layer  & $\mathcal{L}_{CE}(.)$ \\\hline
DFA \citep{Guan2020} &Feature aggregation  & Hint layer  & $\mathcal{L}_{2}(.)$ \\\hline
AdaIN \citep{YangECCV2020} &Feature statistics  & Hint layer  & $\mathcal{L}_{2}(.)$ \\\hline
FN \citep{XuK2020} &Feature representation  & Penultimate layer  & $\mathcal{L}_{CE}(.)$ \\\hline
EC-KD \citep{WangXECCV2020} &Feature representation  & Hint layer  & $\mathcal{L}_{2}(.)$ \\\hline
ALP-KD \citep{Passban2021} &Attention-based layer projection  &Hint layer  &$\mathcal{L}_{2}(.)$ \\\hline
SemCKD \citep{ChenD2021} &Feature maps  &Hint layer  &$\mathcal{L}_{2}(.)$ \\\hline

 \end{tabular}
\end{table*}

The idea of the response-based knowledge is straightforward and easy to understand, especially in the context of ``dark knowledge". From another perspective, the effectiveness of the soft targets is analogous to label smoothing~\citep{Kims2017} or regularizers~\citep{Muller2019,Ding2019}. However, the response-based knowledge usually relies on the output of the last layer, e.g., soft targets, and thus fails to address the intermediate-level supervision from the teacher model, which turns out to be very important for representation learning using very deep neural networks~\citep{Romero2015}. Since the soft logits are in fact  the class probability distribution, the response-based knowledge distillation is also limited to the supervised learning.

\subsection{Feature-Based Knowledge}
\label{3.2}

Deep neural networks are good at learning multiple levels of feature representation with increasing abstraction. This is known as representation learning~\citep{bengio2013representation}. Therefore, both the output of the last layer and the output of intermediate layers, {\it i.e.}, feature maps, can be used as the knowledge to supervise the training of the student model. Specifically, feature-based knowledge from the intermediate layers is a good extension of response-based knowledge, especially for the training of thinner and deeper networks.

The intermediate representations were first introduced in Fitnets~\citep{Romero2015}, to provide hints$\footnote{A hint means the output of a teacher's hidden layer that supervises the student's learning.}$ to improve the training of the student model. The main idea is to directly match the feature activations of the teacher and the student. Inspired by this, a variety of other methods have been proposed to match the features indirectly~\citep{Zagoruyko2017,KimPark2018,Heo2019,Passban2021,ChenD2021,WangXECCV2020}. To be specific, \cite{Zagoruyko2017} derived an ``attention map" from the original feature maps to express knowledge. The attention map was generalized by \cite{HuangW2017} using neuron selectivity transfer. \cite{Passalis2018} transferred knowledge by matching the probability distribution in feature space. To make it easier to transfer the teacher knowledge, \cite{KimPark2018} introduced so called ``factors" as a more understandable form of intermediate representations. To reduce the performance gap between teacher and student, \cite{JinX2019}~proposed route constrained hint learning, which supervises student by outputs of hint layers of teacher. Recently, \cite{Heo2019} proposed to use the activation boundary of the hidden neurons for knowledge transfer. Interestingly, the parameter sharing of intermediate layers of the teacher model together with response-based knowledge is also used as the teacher knowledge~\citep{ZhouFan2018}. To match the semantics between teacher and student, \cite{ChenD2021} proposed cross-layer knowledge distillation, which adaptively assigns proper teacher layers for each student layer via attention allocation.

\begin{figure}[!ht]
\centering
\includegraphics[scale=0.295]{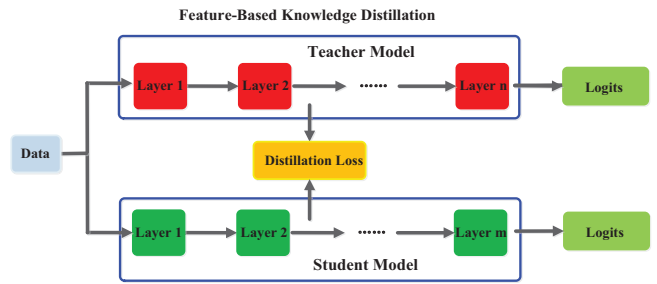}
\caption{The generic feature-based knowledge distillation. }
\label{fig41}
\end{figure}

Generally, the distillation loss for feature-based knowledge transfer can be formulated as
\begin{equation}
\label{FeD}
L_{FeaD}\big(f_{t}(x), f_{s}(x)\big)=\mathcal{L}_{F}\big(\Phi_{t}(f_{t}(x)), \Phi_{s}(f_{s}(x))\big)~,
\end{equation}
where $f_{t}(x)$ and $f_{s}(x)$ are the feature maps of the intermediate layers of teacher and student models, respectively. The transformation functions, $\Phi_{t}(f_{t}(x))$ and $\Phi_{s}(f_{s}(x))$, are usually applied when the feature maps of teacher and student models are not in the same shape. $\mathcal{L}_{F}(.)$ indicates the similarity function used to match the feature maps of teacher and student models. A general feature-based KD model is shown in Fig.~\ref{fig41}. We also summarize different types of feature-based knowledge in Table~\ref{tb1} from the perspective of feature types, source layers, and distillation loss. Specifically, $\mathcal{L}_{2}(.)$, $\mathcal{L}_{1}(.)$, $\mathcal{L}_{CE}(.)$ and $\mathcal{L}_{MMD}(.)$ indicate $l_{2}$-norm distance, $l_{1}$-norm distance, cross-entropy loss and maximum mean discrepancy loss, respectively. Though feature-based knowledge transfer provides favorable information for the learning of the student model, how to effectively choose the hint layers from the teacher model and the guided layers from the student model remains to be further investigated~\citep{Romero2015}. Due to the significant differences between sizes of hint and guided layers, how to properly match feature representations of teacher and student also needs to be explored.

\subsection{Relation-Based Knowledge}
\label{3.3}

Both response-based and feature-based knowledge use the outputs of specific layers in the teacher model. Relation-based knowledge further explores the relationships between different layers or data samples.

To explore the relationships between different feature maps, \cite{Yim2017} proposed a flow of solution process (FSP), which is defined by the Gram matrix between two layers. The FSP matrix summarizes the relations between pairs of feature maps. It is calculated using the inner products between features from two layers. Using the correlations between feature maps as the distilled knowledge, knowledge distillation via singular value decomposition was proposed to extract key information in the feature maps \citep{Lee2018}.
To use the knowledge from multiple teachers, \cite{ZhangPeng2018} formed two graph by respectively using the logits and features of each teacher model as the nodes. Specifically, the importance and relationships of the different teachers are modeled by the logits and representation graphs before the knowledge transfer~\citep{ZhangPeng2018}. Multi-head graph-based knowledge distillation was proposed by \cite{Lee2019}. The graph knowledge is the intra-data relations between any two feature maps via multi-head attention network. To explore the pairwise hint information, the student model also mimics the mutual information flow from pairs of hint layers of the teacher model~\citep{Passalis2020a}. In general, the distillation loss of relation-based knowledge based on the relations of feature maps can be formulated as
\begin{equation}
\label{RelD1}
L_{RelD}(f_{t}, f_{s})=\mathcal{L}_{R^{1}}\big(\Psi_{t}(\hat{f}_{t},\check{f}_{t}), \Psi_{s}(\hat{f}_{s},\check{f}_{s})\big)~,
\end{equation}
where $f_{t}$ and $f_{s}$ are the feature maps of teacher and student models, respectively. Pairs of feature maps are chosen from the teacher model, $\hat{f}_{t}$ and $\check{f}_{t}$, and from the student model, $\hat{f}_{s}$ and $\check{f}_{s}$. $\Psi_{t}(.)$ and $\Psi_{s}(.)$ are the similarity functions for pairs of feature maps from the teacher and student models. $\mathcal{L}_{R^{1}}(.)$ indicates the correlation function between the teacher and student feature maps.

Traditional knowledge transfer methods often involve individual knowledge distillation. The individual soft targets of a teacher are directly distilled into student. In fact, the distilled knowledge contains not only feature information but also mutual relations of data samples \citep{YouS2017,ParkK2019}. Specifically, \cite{LiuC2019} proposed a robust and effective knowledge distillation method via instance relationship graph. The transferred knowledge in instance relationship graph contains instance features, instance relationships and the feature space transformation cross layers. \cite{ParkK2019} proposed a relational knowledge distillation, which transfers the knowledge from instance relations. Based on idea of manifold learning, the student network is learned by feature embedding, which preserves the feature similarities of samples in the intermediate layers of the teacher networks \citep{ChenH2018}. The relations between data samples are modelled as probabilistic distribution using feature representations of data \citep{Passalis2018,Passalis2020}. The probabilistic distributions of teacher and student are matched by knowledge transfer. \citep{Tung2019} proposed a similarity-preserving knowledge distillation method. In particular, similarity-preserving knowledge, which arises from the similar activations of input pairs in the teacher networks, is transferred into the student network, with the pairwise similarities preserved.  \cite{Peng2019} proposed a knowledge distillation method based on correlation congruence, in which the distilled knowledge contains both the instance-level information and the correlations between instances. Using the correlation congruence for distillation, the student network can learn the correlation between instances.

\begin{figure}[!ht]
\centering
\includegraphics[scale=0.37]{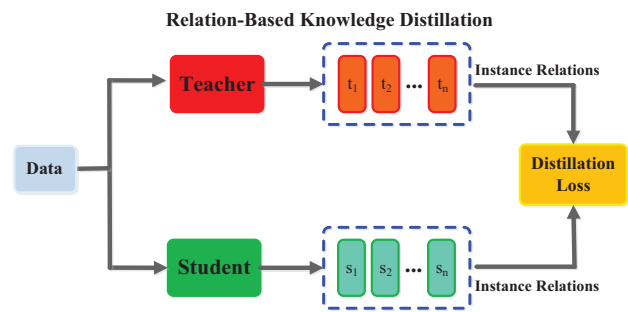}
\caption{The generic instance relation-based knowledge distillation. }
\label{fig42}
\end{figure}

As described above, the distillation loss of relation-based knowledge based on the instance relations can be formulated as
\begin{equation}
\label{RelD2}
L_{RelD}(F_{t}, F_{s})
=\mathcal{L}_{R^{2}}\big(\psi_{t}(t_{i},t_{j}), \psi_{s}(s_{i},s_{j})\big)~,
\end{equation}
where $(t_{i},t_{j})\in F_{t}$ and $(s_{i},s_{j})\in F_{s}$,  and $F_{t}$ and $F_{s}$ are the sets of feature representations from the teacher and student models, respectively. $\psi_{t}(.)$ and $\psi_{s}(.)$ are the similarity functions of $(t_{i},t_{j})$ and $(s_{i},s_{j})$. $\mathcal{L}_{R^{2}}(.)$ is the correlation function between the teacher and student feature representations. A typical instance relation-based KD model is shown in Fig.~\ref{fig42}.

\begin{table*}[!ht]
  \centering
  \caption{A summary of relation-based knowledge}\label{tb2}
\begin{tabular}{|c|c|c|c|}
 \hline
 \multicolumn{4}{|c|}{Relation-based knowledge} \\\hline
Methods    & Knowledge Types   &Knowledge Sources  & Distillation losses \\\hline
FSP \citep{Yim2017} &FSP matrix  &End of multi-layer group & $\mathcal{L}_{2}(.)$ \\\hline
\cite{YouS2017}  &Instance relation  &Hint layers  & $\mathcal{L}_{2}(.)$ \\\hline
\cite{ZhangPeng2018} & Logits graph, Representation graph &Softmax layers, Hint layers & $\mathcal{L}_{EM}(.)$, $\mathcal{L}_{MMD}(.)$ \\\hline
DarkRank \citep{ChenY2018} &Similarity DarkRank & Fully-connected layers & $\mathcal{L}_{KL}(.)$ \\\hline
MHGD \citep{Lee2019}  &Multi-head graph  &Hint layers & $\mathcal{L}_{KL}(.)$ \\\hline
RKD \citep{ParkK2019} &Instance relation  &Fully-connected layers & $\mathcal{L}_{H}(.)$, $\mathcal{L}_{AW}(.)$ \\\hline
IRG \citep{LiuC2019} & Instance relationship graph &Hint layers & $\mathcal{L}_{2}(.)$ \\\hline
SP \citep{Tung2019} &Similarity matrix   &Hint layers  & $\|.\|_{F}$ \\\hline
CCKD \citep{Peng2019}  &Instance relation  &Hint layers  & $\mathcal{L}_{2}(.)$ \\\hline
MLKD \citep{YuYazici2019}  &Instance relation  &Hint layers  & $\|.\|_{F}$  \\\hline
PKT\citep{Passalis2020} &Similarity probability distribution & Fully-connected layers & $\mathcal{L}_{KL}(.)$ \\\hline
\cite{Passalis2020a} & Mutual information flow  & Hint layers & $\mathcal{L}_{KL}(.)$ \\\hline
LP \citep{ChenH2018} &Instance relation  &Hint layers  & $\mathcal{L}_{2}(.)$ \\\hline
 \end{tabular}
\end{table*}

Distilled knowledge can be categorized from different perspectives, such as structured knowledge of the data \citep{LiuC2019,ChenH2018,Peng2019,Tung2019,Tian2020}, privileged information about input features \citep{Lopez2016,Vapnik2015}. A summary of differnet categories of relation-based knowledge is shown in Table~\ref{tb2}. Specifically, $\mathcal{L}_{EM}(.)$, $\mathcal{L}_{H}(.)$, $\mathcal{L}_{AW}(.)$ and $\|.\|_{F}$ are Earth Mover distance, Huber loss, Angle-wise loss and Frobenius norm, respectively. Although some types of relation-based knowledge are provided recently, how to model the relation information from feature maps or data samples as knowledge still deserves further study.

\section{Distillation Schemes}
\label{4}

\begin{figure}[!ht]
\centering
\includegraphics[scale=0.5]{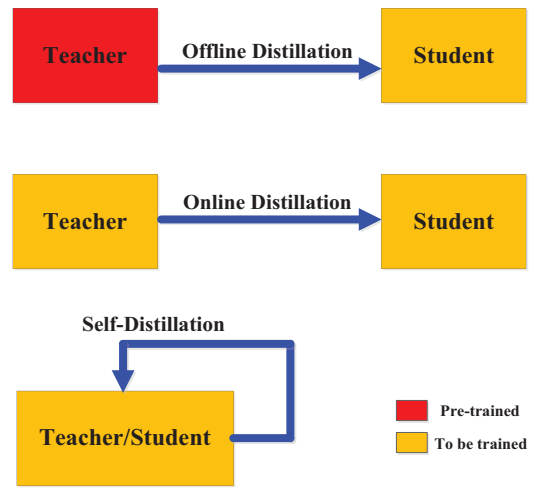}
\caption{Different distillations. The red color for ``pre-trained" means networks are learned before distillation and the yellow color for ``to be trained" means networks are learned during distillation}
\label{fig5}
\end{figure}

In this section, we discuss the distillation schemes ({\it i.e.} training schemes) for both teacher and student models.  According to whether the teacher model is updated simultaneously with the student model or not, the learning schemes of knowledge distillation can be directly divided into three main categories: \textbf{offline distillation}, \textbf{online distillation}  and \textbf{self-distillation}, as shown in Fig.~\ref{fig5}.

\subsection{Offline Distillation}
\label{4.1}
Most of previous knowledge distillation methods work offline. In vanilla knowledge distillation~\citep{Hinton2015}, the knowledge is transferred from a pre-trained teacher model into a student model. Therefore, the whole training process has two stages, namely: 1) the large teacher model is first trained on a set of training samples before distillation; and 2) the teacher model is used to extract the knowledge in the forms of logits or the intermediate features, which are then used to guide the training of the student model during distillation.

The first stage in offline distillation is usually not discussed as part of knowledge distillation, {\it i.e.}, it is assumed that the teacher model is pre-defined. Little attention is paid to the teacher model structure and its relationship with the student model. Therefore, the offline methods mainly focus on improving different parts of the knowledge transfer, including the design of knowledge~\citep{Hinton2015,Romero2015} and the loss functions for matching features or distributions matching~\citep{HuangW2017,Passalis2018,Zagoruyko2017,Seyed2019,LiLi2018,Heob2019,Asif2019}. The main advantage of offline methods is that they are simple and easy to be implemented. For example, the teacher model may contain a set of models trained using different software packages, possibly located on different machines. The knowledge can be extracted and stored in a cache.

The offline distillation methods usually employ one-way knowledge transfer and two-phase training procedure. However, the complex high-capacity teacher model with huge training time can not be avoided, while the training of the student model in offline distillation is usually efficient under the guidance of the teacher model. Moreover, the capacity gap between large teacher and small student always exists, and student often largely relies on teacher.

\subsection{Online Distillation}
\label{4.2}

Although offline distillation methods are simple and effective, some issues in offline distillation have attracted increasing attention from the research community~\citep{Seyed2019}. To overcome the limitation of offline distillation, online distillation is proposed to further improve the performance of the student model, especially when a large-capacity high performance teacher model is not available~\citep{ZhangY2018,Chen2020}. In online distillation, both the teacher model and the student model are updated simultaneously, and the whole knowledge distillation framework is end-to-end trainable.

A variety of online knowledge distillation methods have been proposed, especially in the last few years \citep{ZhangY2018,Chen2020,Xie2019,Anil2018,KimHyun2019,ZhouFan2018,Walawalkar2020,WuG2021,ZhangH2021}. Specifically,  in deep mutual learning~\citep{ZhangY2018}, multiple neural networks work in a collaborative way. Any one network can be the student model and other models can be the teacher during the training process. To improve generalization ability, deep mutual learning is extended by using ensemble of soft logits~\citep{Guo2020}. \cite{Chen2020} further introduced auxiliary peers and a group leader into deep mutual learning to form a diverse set of peer models. To reduce the computational cost, \cite{Lanx2018} proposed a multi-branch architecture, in which each branch indicates a student model and different branches share the same backbone network. Rather than using the ensemble of logits, \cite{KimHyun2019} introduced a feature fusion module to construct the teacher classifier. \cite{Xie2019} replaced the convolution layer with cheap convolution operations to form the student model. \cite{Anil2018} employed online distillation to train large-scale distributed neural network, and proposed a variant of online distillation called co-distillation. Co-distillation in parallel trains multiple models with the same architectures and any one model is trained by transferring the knowledge from the other models. Recently, an online adversarial knowledge distillation method is proposed to simultaneously train multiple networks by the discriminators using knowledge from both the class probabilities and a feature map \citep{OAKD2020ICRL}. Adversarial co-distillation is lately devised by using GAN to generate divergent examples \citep{ZhangH2021}.

Online distillation is a one-phase end-to-end training scheme with efficient parallel computing. However, existing online methods ({\it e.g.}, mutual learning) usually fails to address the high-capacity teacher in online settings, making it an interesting topic to further explore the relationships between the teacher and student model in online settings.

\subsection{Self-Distillation}
\label{4.3}

In self-distillation, the same networks are used for the teacher and the student models \citep{ZhangL2019,HouY2019,Zhangz2020,YangCXie2019,LeeHwang2019,Phuong2019b,Lan2018,XuLiu2019,Mobahi2020}. This can be regarded as a special case of online distillation. Specifically, \cite{ZhangL2019} proposed a new self-distillation method, in which knowledge from the deeper sections of the network is distilled into its shallow sections. Similar to the self-distillation in \citep{ZhangL2019},  a self-attention distillation method was proposed for lane detection~\citep{HouY2019}. The network utilizes the attention maps of its own layers as distillation targets for its lower layers. Snapshot distillation~\citep{YangCXie2019} is a special variant of self-distillation, in which knowledge in the earlier epochs of the network (teacher) is transferred into its later epochs (student) to support a supervised training process within the same network. To further reduce the inference time via the early exit, \cite{Phuong2019b} proposed distillation-based training scheme, in which the early exit layer tries to mimic the output of later exit layer during the training. Recently, self-distillation has been theoretically analyzed in~\citep{Mobahi2020}, and its improved performance experimentally demonstrated in~\citep{Zhangz2020}.

Furthermore, some interesting self-distillation methods are recently proposed~\citep{YuanL2019,Sukmin2019,Hahn2019}. To be specific, Yuan {\it et al.} proposed  teacher-free knowledge distillation methods based on the analysis of label smoothing regularization~\citep{YuanL2019}. Hahn and  Choi proposed a novel self-knowledge distillation method, in which the self-knowledge consists of the predicted probabilities instead of traditional soft probabilities \citep{Hahn2019}. These predicted probabilities are defined by the feature representations of the training model. They reflect the similarities of data in feature embedding space. Yun {\it et al.} proposed class-wise self-knowledge distillation to match the output distributions of the training model between intra-class samples and augmented samples within the same source with the same model~\citep{Sukmin2019}. In addition, the self-distillation proposed by \cite{LeeHwang2019} is adopted for data augmentation and the self-knowledge of augmentation is distilled into the model itself. Self distillation is also adopted to optimize deep models (the teacher or student networks) with the same architecture one by one~\citep{Furlanello2018,Bagherinezhad2018}. Each network distills the knowledge of the previous network using a teacher-student optimization.


Besides, offline, online and self distillation can also be intuitively understood from the perspective of human beings teacher-student learning. Offline distillation means the knowledgeable teacher teaches a student knowledge; online distillation means both teacher and student study together with each other; self-distillation means student learn knowledge by oneself. Moreover, just like the human beings learning, these three kinds of distillation can be combined to complement each other due to their own advantages. For example, both self-distillation and online distillation are properly integrated via the multiple knowledge transfer framework~\citep{SunL2021}.

\section{Teacher-Student Architecture}
\label{5}

\begin{figure}[!ht]
\centering
\includegraphics[scale=0.47]{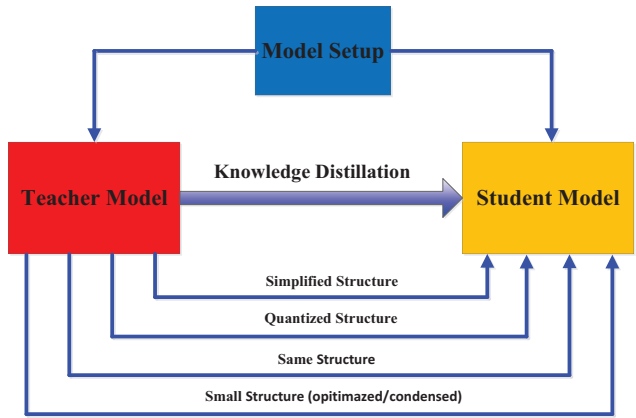}
\caption{Relationship of the teacher and student models. }
\label{fig6}
\end{figure}

In knowledge distillation, the teacher-student architecture is a generic carrier to form the knowledge transfer. In other words, the quality of knowledge acquisition and distillation from teacher to student is also determined by how to design the teacher and student networks. In terms of the habits of human beings learning, we hope that a student can find a right teacher. Thus, to well finish capturing and distilling knowledge in knowledge distillation, how to select or design proper structures of teacher and student is very important but difficult problem. Recently, the model setups of teacher and student are almost pre-fixed with unvaried sizes and structures during distillation, so as to easily cause the model capacity gap. However, how to particulary design the architectures of teacher and student and why their architectures are determined by these model setups are nearly missing. In this section, we discuss the relationship between the structures of the teacher model and the student model as illustrated in Fig.~\ref{fig6}.

Knowledge distillation was previously designed to compress an ensemble of deep neural networks in \citep{Hinton2015}. The complexity of deep neural networks mainly comes from two dimensions: depth and width. It is usually required to transfer knowledge from deeper and wider neural networks to  shallower and thinner neural networks~\citep{Romero2015}. The student network is usually chosen to be: 1) a simplified version of a teacher network with fewer layers and fewer channels in each layer~\citep{WangH2018,Lanx2018,LiLi2018}; or 2) a quantized version of a teacher network in which the structure of the network is preserved~\citep{Polino2018,Mishra2018,WeiY2018,Shin2018}; or 3) a small network with efficient basic operations~\citep{howard2017mobilenets,ZhangZhou2018,HuangLiu2017}; or 4) a small network with optimized global network structure~\citep{LiuJia2019,xie2019selftraining,GuJ2020}; or 5) the same network as teacher \citep{ZhangY2018,Furlanello2018,Tarvainen2017}.

The model capacity gap between the large deep neural network and a small student neural network can degrade knowledge transfer~\citep{Seyed2019,GaoM2021}. To effectively transfer knowledge to student networks, a variety of methods have been proposed for a controlled reduction of the model complexity \citep{ZhangY2018,Nowak2018,Crowley2018,LiuPeng2019,LiuJia2019,WangH2018,GuJ2020}. Specifically, \cite{Seyed2019} introduced a teacher assistant to mitigate the training gap between teacher model and student model. The gap is further reduced by residual learning, {\it i.e.}, the assistant structure is used to learn the residual error~\citep{GaoM2021}. On the other hand, several recent methods also focus on minimizing the difference in structure of the student model and the teacher model. For example, \cite{Polino2018} combined network quantization with knowledge distillation, {\it i.e.}, the student model is small and quantized version of the teacher model. \cite{Nowak2018} proposed a structure compression method which involves transferring the knowledge learned by multiple layers to a single layer. \cite{WangH2018} progressively performed block-wise knowledge transfer from teacher networks to student networks while preserving the receptive field. In online setting, the teacher networks are usually ensembles of student networks, in which the student models share similar structure (or the same structure) with each other~\citep{ZhangY2018,Lanx2018,Furlanello2018,Chen2020}.

Recently, depth-wise separable convolution has been widely used to design efficient neural networks for mobile or embedded devices \citep{Chollet2017,howard2017mobilenets,sandler2018mobilenetv2,ZhangZhou2018,ma2018shufflenet}.
Inspired by the success of neural architecture search (or NAS), the performances of small neural networks have been further improved by searching for a global structure based on efficient meta operations or blocks~\citep{Wu2019CVPR,Tan2019CVPR,tan19a,radosavovic2020designing}. Furthermore, the idea of dynamically searching for a knowledge transfer regime also appears in knowledge distillation, {\it e.g.}, automatically removing redundant layers in a data-driven way using reinforcement learning~\citep{Ashok2018},  and searching for optimal student networks given the teacher networks~\citep{LiuJia2019,xie2019selftraining,GuJ2020}.

Most previous works focus on designing either the structures of teacher and student models or the knowledge transfer scheme between them. To make a small student model well match a large teacher model for improving knowledge distillation performance, the adaptive teacher-student learning architecture is necessary. Recently, the idea of a neural architecture search in knowledge distillation, {\it i.e.}, a joint search of student structure and knowledge transfer under the guidance of the teacher model, will be an interesting subject of future study.


\section{Distillation Algorithms}
\label{6}

A simple yet very effective idea for knowledge transfer is to directly match the response-based knowledge, feature-based knowledge~\citep{Romero2015, Hinton2015} or the representation distributions in feature space~\citep{Passalis2018} between the teacher model and the student model. Many different algorithms have been proposed to improve the process of transferring knowledge in more complex settings. In this section, we review recently proposed typical types of distillation methods for knowledge transfer within the field of knowledge distillation.

\subsection{Adversarial Distillation}
\label{6.1}
In knowledge distillation, it is difficult for the teacher model to perfectly learn from the true data distribution. Simultaneously, the student model has only a small capacity and so cannot mimic the teacher model accurately~\citep{Seyed2019}. Are there other ways of training the student model in order to mimic the teacher model? Recently, adversarial learning has received a great deal of attention due to its great success in generative networks, {\it i.e.}, generative adversarial networks or GANs~\citep{goodfellow2014}. Specifically, the discriminator in a GAN estimates the probability that a sample comes from the training data distribution while the generator tries to fool the discriminator using generated data samples. Inspired by this, many adversarial knowledge distillation methods have been proposed to enable the teacher and student networks to have a better understanding of the true data distribution \citep{WangXZ2018,ZhengXu2018a,Micaelli2019,ZhengXu2018b,LiuFusi2018,WangTao2019,ChenH2019,ShenHe2019,Changyong2019,LiuP2018,Bela2018}.

\begin{figure}[!ht]
\centering
\includegraphics[scale=0.3]{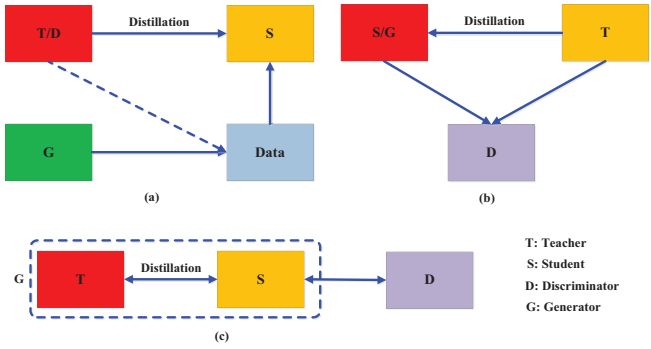}
\caption{The different categories of the main adversarial distillation  methods. (a) Generator in GAN produces training data to improve KD performance; the teacher may be used as discriminator. (b) Discriminator in GAN ensures that the student (also as generator) mimics the teacher. (c) Teacher and student form a generator; online knowledge distillation is enhanced by the discriminator. }
\label{fig7}
\end{figure}

As shown in Fig.~\ref{fig7},  adversarial learning-based distillation methods, especially those methods using GANs, can be divided into three main categories as follows. In the first category, an adversarial generator is trained to generate synthetic data, which is either directly used as the training dataset~\citep{ChenH2019,YeJ2020} or used to augment the training dataset~\citep{LiuFusi2018}, shown in Fig.~\ref{fig7}~(a). Furthermore, \cite{Micaelli2019} utilized an adversarial generator to generate hard examples for knowledge transfer.  Generally, the distillation loss used in this GAN-based KD category can be formulated as
\begin{equation}
\label{GD1}
L_{KD}=\mathcal{L}_{G}\big(F_{t}(G(z)), F_{s}(G(z))\big)~,
\end{equation}
where $F_{t}(.)$ and $F_{s}(.)$ are the outputs of the teacher and student models, respectively. $G(z)$ indicates the training samples generated by the generator $G$ given the random input vector $z$, and $\mathcal{L}_{G}$ is a distillation loss to force the match between the predicted and the ground-truth probability distributions, e.g., the cross entropy loss or the Kullback-Leibler (KL) divergence loss.

To make student well match teacher, a discriminator in the second category is introduced to distinguish the samples from the student and the teacher models by using either the logits~\citep{ZhengXu2018a,ZhengXu2018b} or the features~\citep{WangTao2019}, shown in Fig.~\ref{fig7}~(b). Specifically, \cite{Bela2018} used unlabeled data samples to form the knowledge transfer. Multiple discriminators were used by~\cite{ShenHe2019}. Furthermore, an effective intermediate supervision, {\it i.e.}, the squeezed knowledge, was used by \cite{Changyong2019} to mitigate the capacity gap between the teacher and the student.  A representative model proposed by \cite{WangTao2019} falls into this category, which can be formulated as
\begin{equation}
\label{GD2}
\begin{aligned}
L_{GANKD}=&\mathcal{L}_{CE}\big(G(F_{s}(x)),y\big)+\alpha\mathcal{L}_{KL}\big(G(F_{s}(x)),F_{t}(x)\big)\\
&~~~~~+\beta\mathcal{L}_{GAN}\big(F_{s}(x), F_{t}(x)\big)~,
\end{aligned}
\end{equation}
where $G$ is a student network and $\mathcal{L}_{GAN}(.)$ indicates a typical loss function used in generative adversarial network to make the outputs between student and teacher as similar as possible.

In the third category, adversarial knowledge distillation is carried out in an online manner, {\it i.e.}, the teacher and the student are jointly optimized in each iteration~\citep{WangXZ2018,OAKD2020ICRL}, shown in Fig.~\ref{fig7}~(c). Besides, using knowledge distillation to compress GANs, a learned small GAN student network mimics a larger GAN teacher network via knowledge transfer \citep{Aguinaldo2019,LiM2020}.

In summary, three main points can be concluded from the adversarial distillation methods above as follows: GAN is an effective tool to enhance the power of student learning via the teacher knowledge transfer; joint GAN and KD can generate the valuable data for improving the KD performance and overcoming the limitations of unusable and unaccessible data; KD can be used to compress GANs.

\subsection{Multi-Teacher Distillation }
\label{6.2}



\begin{figure}[!ht]
\centering
\includegraphics[scale=0.39]{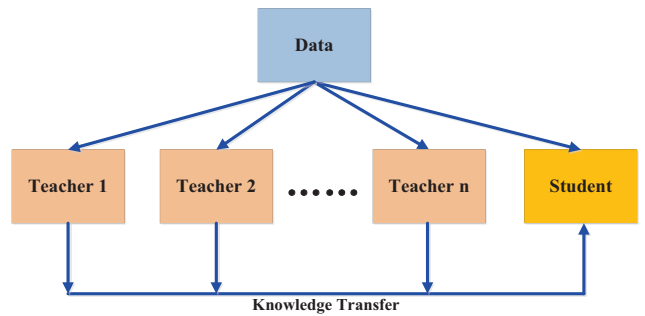}
\caption{The generic framework for multi-teacher distillation. }
\label{fig8}
\end{figure}

Different teacher architectures can provide their own useful knowledge for a student network. The multiple teacher networks can be individually and integrally used for distillation during the period of training a student network. In a typical teacher-student framework, the teacher usually has a large model or an ensemble of large models. To transfer knowledge from multiple teachers, the simplest way is to use the averaged response from all teachers as the supervision signal~\citep{Hinton2015}. Several multi-teacher knowledge distillation methods have recently been proposed \citep{Sau2016,YouS2017,Chensu2019,Furlanello2018,YangC2019,ZhangY2018,LeeKw2020,Park2019,Papernot2017,Fukuda2017,Ruder2017,WuA2019,YangZ2020,Vongku2019,ZhaoH2020,YuanF2021}.
A generic framework for multi-teacher distillation is shown in Fig.~\ref{fig8}.

Multiple teacher networks have turned out to be effective for training student model usually using logits and feature representation as the knowledge. In addition to the averaged logits from all teachers, \cite{YouS2017} further incorporated features from the intermediate layers in order to encourage the dissimilarity among different training samples. To utilize both logits and intermediate features, \cite{Chensu2019} used two teacher networks, in which one teacher transfers response-based knowledge to the student and the other teacher transfers feature-based knowledge to the student. \cite{Fukuda2017} randomly selected one teacher from the pool of teacher networks at each iteration. To transfer feature-based knowledge from multiple teachers, additional teacher branches are added to the student networks to mimic the intermediate features of teachers~\citep{Park2019,Asif2019}. Born again networks address multiple teachers in a step-by-step manner, {\it i.e.}, the student at the $t$ step is used as the teacher of the student at the ${t+1}$ step~\citep{Furlanello2018}, and similar ideas can be found in \citep{YangC2019}. To efficiently perform knowledge transfer and explore the power of multiple teachers, several alternative methods have been proposed to simulate multiple teachers by adding different types of noise to a given teacher~\citep{Sau2016} or by using stochastic blocks and skip connections~\citep{LeeKw2020}. Using multiple teacher models with feature ensembles, knowledge amalgamation is designed in \citep{ShenCW2019,LuoS2019,ShenCX2019,LuoECCV2020}. Through knowledge amalgamation, many public available trained deep models as teachers can be reused. More interestingly, due to the special characteristics of multi-teacher distillation, its extensions are used for domain adaptation via knowledge adaptation~\citep{Ruder2017}, and to protect the privacy and security of data \citep{Vongku2019,Papernot2017}.

\begin{table}[!ht]
  \centering
  \caption{A summary of multi-teacher distillation using different types of knowledge and distillation schemes. The response-based knowledge, feature-based knowledge and relation-based knowledge are abbreviated as `ResK', `FeaK' and `RelK', respectively.}\label{mt}
\begin{tabular}{|c|c|c|c|}
 \hline
 \multicolumn{4}{|c|}{\textbf{Offline Distillation}}\\\hline
 Methods           & ResK      & FeaK   & RelK     \\\hline
\cite{YouS2017}  &\CheckmarkBold  & \XSolidBold    &\CheckmarkBold  \\\hline
\cite{Fukuda2017} &\CheckmarkBold   &\XSolidBold   &\XSolidBold   \\\hline
\cite{ShenCX2019} &\CheckmarkBold  & \CheckmarkBold   &\XSolidBold  \\\hline
\cite{WuA2019} &\XSolidBold  &\XSolidBold   &\CheckmarkBold     \\\hline
\cite{Park2019} &\XSolidBold   & \CheckmarkBold  &\XSolidBold    \\\hline
\cite{YangZ2020} &\CheckmarkBold   &\XSolidBold   &\XSolidBold   \\\hline
\cite{LuoECCV2020} &\CheckmarkBold   &\CheckmarkBold   &\XSolidBold   \\\hline
\cite{Kwon2020} &\CheckmarkBold   &\XSolidBold   &\XSolidBold   \\\hline
\cite{LiuY2020} &\CheckmarkBold   &\CheckmarkBold   &\XSolidBold   \\\hline
\cite{ZhaoH2020} &\CheckmarkBold   &\CheckmarkBold   &\XSolidBold   \\\hline
\cite{YuanF2021} &\CheckmarkBold   &\XSolidBold   &\XSolidBold   \\\hline
 \multicolumn{4}{|c|}{\textbf{Online Distillation}}\\\hline
 Methods           & ResK      & FeaK   & RelK     \\\hline
 \cite{Papernot2017} &\CheckmarkBold   &\XSolidBold   &\XSolidBold  \\\hline
\cite{Furlanello2018} &\CheckmarkBold  & \XSolidBold  &\XSolidBold  \\\hline
\cite{ZhangY2018} &\CheckmarkBold  & \XSolidBold  &\XSolidBold    \\\hline
\cite{YangC2019} &\CheckmarkBold  & \XSolidBold  &\XSolidBold   \\\hline
\cite{LeeKw2020} &\CheckmarkBold  & \CheckmarkBold  &\XSolidBold   \\\hline

 \end{tabular}
\end{table}

A summary of typical multi-teacher distillation methods using different types of knowledge and distillation schemes is shown in Table \ref{mt}. Generally, multi-teacher knowledge distillation can provide rich knowledge and tailor a versatile student model because of the diverse knowledge from different teachers. However, how to effectively integrate different types of knowledge from multiple teachers needs to be further studied.

\subsection{Cross-Modal Distillation}
\label{6.3}

The data or labels for some modalities might not be available during training or testing~\citep{Gupta2016,Garcia2018,ZhaoLi2018,Roheda2018,ZhangLP2020}.  For this reason it is important to transfer knowledge between different modalities. Several typical scenarios using cross-modal knowledge transfer are reviewed as follows.

\begin{figure}[!ht]
\centering
\includegraphics[scale=0.42]{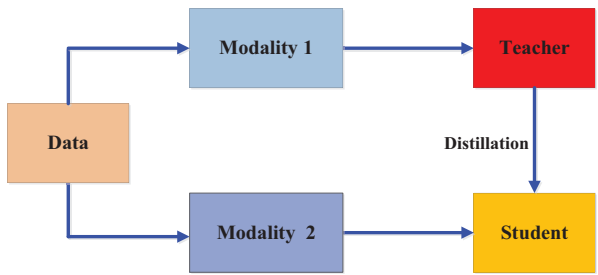}
\caption{The generic framework for cross-modal distillation. For simplicity,  only two modalities are shown.}
\label{fig9}
\end{figure}

Given a teacher model pretrained on one modality (e.g., RGB images) with a large number of well-annotated data samples, \cite{Gupta2016} transferred the knowledge from the teacher model to the student model with a new unlabeled input modality, such as a depth image and optical flow. Specifically, the proposed method relies on unlabeled paired samples involving both modalities, {\it i.e.}, both RGB and depth images. The features obtained from RGB images by the teacher are then used for the supervised training of the student~\citep{Gupta2016}. The idea behind the paired samples is to transfer the annotation or label information via pair-wise sample registration and has been widely used for cross-modal applications~\citep{Albanie2018,ZhaoLi2018,Thoker2019}. To perform human pose estimation through walls or with occluded images, \cite{ZhaoLi2018} used synchronized radio signals and camera images. Knowledge is transferred across modalities for radio-based human pose estimation. \cite{Thoker2019} obtained paired samples from two modalities: RGB videos and skeleton sequence. The pairs are used to transfer the knowledge learned on RGB videos to a skeleton-based human action recognition model. To improve the action recognition performance using only RGB images, \cite{Garcia2018} performed cross-modality distillation on an additional modality, {\it i.e.}, depth image, to generate a hallucination stream for RGB image modality. \cite{Tian2020} introduced a contrastive loss to transfer pair-wise relationship across different modalities. To improve target detection, \cite{Roheda2018} proposed cross-modality distillation among the missing and available modelities using GANs. The generic framework of cross-modal distillation is shown in Fig.~\ref{fig9}.

\begin{table*}[!ht]
  \centering
  \caption{A summary of cross-modal distillation with modalities, types of knowledge and distillation.}\label{mcd}
\begin{tabular}{|c|m{4cm}<{\centering}|c|c|c|}
 \hline
Methods           &Modality for Teacher      & Modality for Student   & Knowledge    &Distillation   \\\hline
\cite{Hoffman2016} &RGB images  & Depth images    &FeaK  &Offline \\\hline
\cite{Gupta2016} &RGB images  & Depth images    &ResK  &Offline \\\hline
\cite{Passalis2018} &Textual modality & Visual modality   &RelK  &Offline \\\hline
\cite{Garcia2018} &Depth and RGB videos  & RGB videos    &ResK, FeaK  &Offline \\\hline
\cite{ZhaoLi2018} &RGB frames  & Radio frequency heatmaps    &ResK &Offline \\\hline
\cite{Roheda2018} &Temporal data  & Spatial data   &FeaK  &Online \\\hline
\cite{Albanie2018} &Vision  & Sound   &ResK  &Offline \\\hline
\cite{Thoker2019} &RGB videos  & Skeleton data   &ResK  &Offline \\\hline
\cite{DoT2019} &Images, question, answer
information  & Image-questions   &ResK  &Offline \\\hline
\cite{Tian2020} &RGB images  & Depth images   &ResK  &Offline \\\hline
\cite{GaoZ2020} & Multi-modal images    &Single-mode images   & ResK, FeaK   &Offline \\\hline
 \end{tabular}
\end{table*}

Moreover, \cite{DoT2019} proposed a knowledge distillation-based visual question answering method, in which knowledge from trilinear interaction teacher model with image-question-answer as inputs is distilled into the learning of a bilinear interaction student model with image-question as inputs. The probabilistic knowledge distillation proposed by \cite{Passalis2018} is also used for knowledge transfer from the textual modality into the visual modality. \cite{Hoffman2016} proposed a modality hallucination architecture based on cross-modality distillation to improve detection performance. Besides, these cross-model distillation methods also transfer the knowledge among multiple domains \citep{Kundu2019,ChenY2019,SuJ2016}.


A summary of cross-modal distillation with different modalities, types of knowledge and distillation schemes is shown in Table~\ref{mcd}. Specifically, it can be seen that knowledge distillation performs well in visual recognition tasks in the cross-modal scenarios. However, cross-modal knowledge transfer is a challenging study when there is a modality gap, e.g., lacking of the paired samples between different modalities.

\subsection{Graph-Based Distillation}
\label{6.4}

\begin{figure}[!ht]
\centering
\includegraphics[scale=0.45]{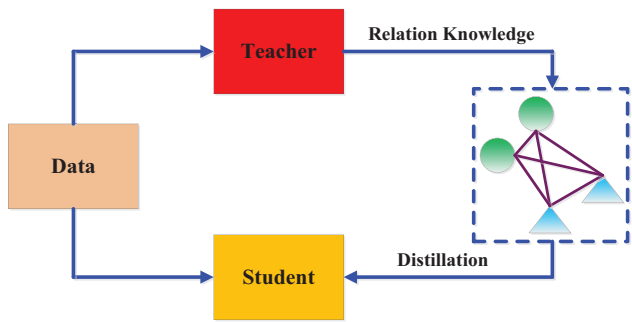}
\caption{A generic framework for graph-based distillation. }
\label{fig10}
\end{figure}

Most of knowledge distillation algorithms focus on transferring individual instance knowledge from the teacher to the student, while some recent methods have been proposed to explore the intra-data relationships using graphs~\citep{ChenH2018,ZhangPeng2018,Lee2019,ParkK2019,yao2019graph,ma2019graph,HouY2020}. The main ideas of these graph-based distillation methods are 1) to use the graph as the carrier of teacher knowledge; or 2) to use the graph to control the message passing of the teacher knowledge. A generic framework for graph-based distillation is shown in Fig.~\ref{fig10}. As described in Section~\ref{3.3}, the graph-based knowledge falls in line of relation-based knowledge. In this section, we introduce typical definitions of the graph-based knowledge and the graph-based message passing distillation algorithms.

Specifically, in \citep{ZhangPeng2018}, each vertex represents a self-supervised teacher. Two graphs are then constructed using logits and intermediate features, {\it i.e.}, the logits graph and representation graph, to transfer knowledge from multiple self-supervised teachers to the student. In \citep{ChenH2018}, the graph is used to maintain the relationship between samples in the high-dimensional space. Knowledge transfer is then carried out using a proposed locality preserving loss function. \cite{Lee2019} analysed intra-data relations using a multi-head graph, in which the vertices are the features from different layers in CNNs. \cite{ParkK2019} directly transferred the mutual relations of data samples, {\it i.e.}, to match edges between a teacher graph and a student graph. \cite{Tung2019} used the similarity matrix to represent the mutual relations of the activations of the input pairs in teacher and student models. The similarity matrix of student matches that of teacher. Furthermore, \cite{Peng2019} not only matched the response-based and feature-based knowledge, but also used the graph-based knowledge. In \citep{LiuC2019}, the instance features and instance relationships are modeled as vertexes and edges of the graph, respectively.

Rather than using the graph-based knowledge, several methods control knowledge transfer using a graph. Specifically, \cite{LuoZ2018} considered the modality discrepancy to incorporate privileged information from the source domain.  A directed graph, referred to as a distillation graph is introduced to explore the relationship between different modalities. Each vertex represent a modality and the edges indicate the connection strength between one modality and another. \cite{Minami2019} proposed a bidirectional graph-based diverse collaborative learning to explore diverse knowledge transfer patterns. \cite{yao2019graph} introduced GNNs to deal with the knowledge transfer for graph-based knowledge. Besides, using knowledge distillation, the topological semantics of a graph convolutional teacher network as the topology-aware knowledge are transferred into the graph convolutional student network~\citep{YangQ2020}

Graph-based distillation can transfer the informative structure knowledge of data. However, how to properly construct graph to model the structure knowledge of data is a still challenging study.

\subsection{Attention-Based Distillation}
Since attention can well reflect the neuron activations of convolutional neural network, some attention mechanisms are used in knowledge distillation to improve the performance of the student network \citep{Zagoruyko2017,HuangW2017,Srinivas2018,Crowley2018,SongX2018}. Among these attention-based KD methods \citep{Crowley2018,HuangW2017,Srinivas2018, Zagoruyko2017}, different attention transfer mechanisms are defined for distilling knowledge from the teacher network to the student network. The core of attention transfer is to define the attention maps for feature embedding in the layers of a neural network. That is to say, knowledge about feature embedding is transferred using attention map functions. Unlike the attention maps, a different attentive knowledge distillation method was proposed by \cite{SongX2018}. An attention mechanism is used to assign different confidence rules \citep{SongX2018}.

\subsection{Data-Free Distillation}

Some data-free KD methods have been proposed to overcome problems with unavailable data arising from privacy, legality, security and confidentiality concerns~\citep{ChenH2019,Raphael2017,Nayak2019,Micaelli2019,Haroush2020,YeJ2020,Nayak2021,Chawla2021}. Just as ``data free"  implies, there is no training data. Instead, the data is newly or synthetically generated.

\begin{figure}[!ht]
\centering
\includegraphics[scale=0.52]{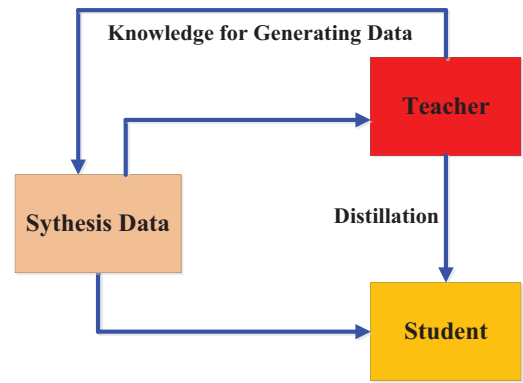}
\caption{A generic framework for data-free distillation. }
\label{fig13}
\end{figure}

Specifically, in \citep{ChenH2019,YeJ2020,Micaelli2019,YooJ2019,HuH2020}, the transfer data is generated by a GAN. In the proposed data-free knowledge distillation method \citep{Raphael2017}, the transfer data to train the student network is reconstructed by using the layer activations or layer spectral activations of the teacher network. \cite{YinH2020} proposed DeepInversion, which uses knowledge distillation to generate synthesized images for data-free knowledge transfer. \cite{Nayak2019} proposed  zero-shot knowledge distillation that does not use existing data. The transfer data is produced by modelling the softmax space using the parameters of the teacher network. In fact, the target data in \citep{Micaelli2019,Nayak2019} is generated by using the information from the feature representations of teacher networks. Similar to zero-shot learning, a knowledge distillation method with few-shot learning is designed by distilling knowledge from a teacher model into a student neural network~\citep{Kimura2018,ShenC2021}. The teacher uses limited labelled data. Besides, there is a new type of distillation called data distillation, which is similar to data-free distillation \citep{Radosavovic2018,LiuPK2019,ZhangW2020}. In data distillation, new training annotations of unlabeled data generated from the teacher model are employed to train a student model.

In summary, the synthesis data in data-free distillation is usually generated from the feature representations from the pre-trained teacher model, as shown in Fig.~\ref{fig13}. Although the data-free distillation has shown a great potential under the condition of unavailable data, it remains a very challenging task, i.e., how to generate high quality diverse training data to improve the model generalizability.

\subsection{Quantized Distillation}

\begin{figure}[!ht]
\centering
\includegraphics[scale=0.48]{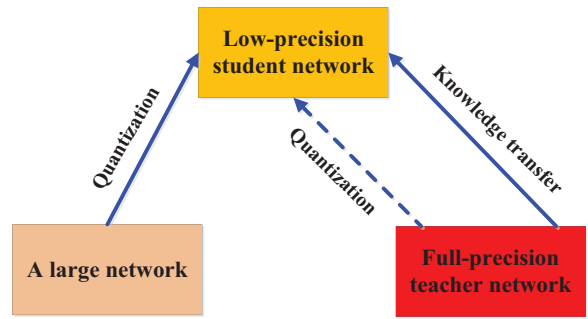}
\caption{A generic framework for quantized distillation. }
\label{fig12}
\end{figure}

Network quantization reduces the computation complexity of neural networks by converting high-precision networks ({\it e.g.}, 32-bit floating point) into low-precision networks ({\it e.g.}, 2-bit and 8-bit). Meanwhile, knowledge distillation aims to train a small model to yield a performance comparable to that of a complex model. Some KD methods have been proposed  using the quantization process in the teacher-student framework \citep{Polino2018,Mishra2018,WeiY2018,Shin2018,KimBha2019}. A  framework for quantized distillation methods is shown in Fig.~\ref{fig12}.

Specifically, \cite{Polino2018} proposed a quantized distillation method to transfer the knowledge to a weight-quantized student network. In \citep{Mishra2018}, the proposed quantized KD is called the ``apprentice". A high precision teacher network transfers knowledge to a small low-precision student network. To ensure that a small student network accurately mimics a large teacher network, the full-precision teacher network is first quantized on the feature maps, and then the knowledge is transferred from the quantized teacher to a quantized student network~\citep{WeiY2018}. \cite{KimBha2019} proposed quantization-aware knowledge distillation, which is based on self-study of a quantized student network and on the co-studying of teacher and student networks with knowledge transfer. Furthermore, \cite{Shin2018} carried out empirical analysis of deep neural networks using both distillation and quantization, taking into account the hyper-parameters for knowledge distillation, such as the size of teacher networks and the distillation temperature. Recently, unlike the quantized distillation methods above, a self-distillation training schemes is designed to improve the performance of quantized deep models, where teacher shares model parameters of student~\citep{BooY2021}.

\subsection{Lifelong Distillation}

Lifelong learning, including continual learning, continuous learning and meta-learning, aims to learn in a similar way to human. It accumulates the previously learned knowledge and also transfers the learned knowledge into future learning \citep{BingLiu2019}. Knowledge distillation provides an effective way to preserve and transfer learned knowledge without catastrophic forgetting. Recently, an increasing number of KD variants, which are based on lifelong learning, have been developed~\citep{Jang2019,Flennerhag2019,PengZ2019,QingLiu2019,LeeKK2019,Zhai2019,ZhouP2019,Shmelkov2017,LiZ2018,Caccia2020}. The methods proposed in \citep{Jang2019,PengZ2019,QingLiu2019,Flennerhag2019} adopt meta-learning. \cite{Jang2019} designed meta-transfer networks that can determine what and where to transfer in the teacher-student architecture.  \cite{Flennerhag2019} proposed a light-weight framework called Leap for meta-learning over task manifolds by transferring knowledge from one learning process to another. \cite{PengZ2019} designed a new knowledge transfer network architecture for few-shot image recognition. The architecture simultaneously incorporates visual information from images and prior knowledge. \cite{QingLiu2019} proposed the semantic-aware knowledge preservation method for image retrieval. The teacher knowledge obtained from the image modalities and semantic information are preserved and transferred.

Moreover, to address the problem of catastrophic forgetting in lifelong learning, global distillation \citep{LeeKK2019}, knowledge distillation-based lifelong GAN \citep{Zhai2019}, multi-model distillation \citep{ZhouP2019} and the other KD-based methods~\citep{LiZ2018,Shmelkov2017} have been developed to extract the learned knowledge and teach the student network on new tasks.

\begin{table*}[!ht]
  \centering
  \caption{Performance comparison of different knowledge distillation methods on CIFAR10. Note that  $\uparrow$ indicates the performance improvement of the student network learned by each method comparing with the corresponding baseline model.}\label{tbp10}
\begin{tabular}{|c|c|c|c|c|}
 \hline
\multicolumn{5}{|c|}{\textbf{Offline Distillation}} \\\hline
Methods               & Knowledge  & Teacher (baseline)  &Student (baseline)  &Accuracies  \\\hline
FSP \citep{Yim2017}   & RelK       &ResNet26 (91.91)  &ResNet8 (87.91)        &88.70       (0.79 $\uparrow$) \\\hline
FT \citep{KimPark2018} & FeaK      &ResNet56 (93.61)  &ResNet20 (92.22)       &93.15    (0.93 $\uparrow$)\\\hline
IRG \citep{LiuC2019}  & RelK       &ResNet20 (91.45)  &ResNet20-x0.5 (88.36)  &90.69     (2.33 $\uparrow$)\\\hline
SP \citep{Tung2019}   &RelK        &WRN-40-1 (93.49)  &WRN-16-1 (91.26)       &91.87       (0.61 $\uparrow$)\\\hline
SP \citep{Tung2019}   &RelK        &WRN-40-2 (95.76)  &WRN-16-8 (94.82)       &95.45       (0.63 $\uparrow$)\\\hline
FN \citep{XuK2020}   &FeaK         &ResNet110 (94.29) &ResNet56 (93.63)       &94.14       (0.51 $\uparrow$)\\\hline
FN \citep{XuK2020}   &FeaK         &ResNet56 (93.63)  &ResNet20 (92.11)       &92.67       (0.56 $\uparrow$)\\\hline
AdaIN \citep{YangECCV2020}  &FeaK         &ResNet26 (93.58)  &ResNet8 (87.78)        &89.02      (1.24 $\uparrow$)\\\hline
AdaIN \citep{YangECCV2020}  &FeaK         &WRN-40-2 (95.07)  &WRN-16-2 (93.98)        &94.67   (0.69 $\uparrow$)\\\hline
AE-KD \citep{Du2020}       &FeaK        &ResNet56 (---)  &MobileNetV2 (75.97)        &77.07       (1.10 $\uparrow$)\\\hline
JointRD \citep{LiG2020}  &FeaK        &ResNet34 (95.39)      &plain-CNN 34 (93.73)   &94.78       (1.05 $\uparrow$)\\\hline
TOFD \citep{ZhangL2020}  &FeaK        &ResNet152 (---)      &ResNeXt50-4  (94.49)   &97.09       (2.60 $\uparrow$)\\\hline
TOFD \citep{ZhangL2020}  &FeaK        &ResNet152 (---)      &MobileNetV2  (90.43)   &93.34       (2.91 $\uparrow$)\\\hline
CTKD \citep{ZhaoH2020}  &RelK, FeaK   &WRN-40-1 (93.43)      &WRN-16-1  (91.28)   &92.50   (1.22 $\uparrow$)\\\hline
CTKD \citep{ZhaoH2020}  &RelK, FeaK   &WRN-40-2 (94.70)      &WRN-16-2  (93.68)   &94.42    (0.74 $\uparrow$)\\\hline
\multicolumn{5}{|c|}{\textbf{Online Distillation}} \\\hline
Methods                        & Knowledge  & Teacher (baseline)     &Student (baseline)                &Accuracies   \\\hline
Rocket-KD \citep{ZhouFan2018}  & FeaK      &WRN-40-1 (93.42)   &WRN-16-1 (91.23)       &92.48  (1.25 $\uparrow$)\\\hline
DML \citep{ZhangY2018}         & ResK      &WRN-28-10 (95.01)  &ResNet32 (92.47)       &95.75, 93.18  (0.71 $\uparrow$)\\\hline
DML \citep{ZhangY2018}         & ResK     &MobileNet (93.59)   &ResNet32 (92.47)       &94.24, 93.32  (0.85 $\uparrow$)\\\hline
DML \citep{ZhangY2018}         & ResK     &ResNet32 (92.47)    &ResNet32 (92.47)       &92.68, 92.80  (0.33 $\uparrow$)\\\hline
ONE \citep{Lanx2018}           &ResK      &ResNet32+ONE        &ResNet32 (93.07)       &94.01        (0.84 $\uparrow$)\\\hline
ONE \citep{Lanx2018}           &ResK      &ResNet110+ONE       &ResNet110 (94.44)      &94.83        (0.39 $\uparrow$)\\\hline
PCL \citep{WuG2021}  &ResK    &Student ensemble         &ResNet110 (94.91)        &95.53   (0.62 $\uparrow$)\\\hline
PCL \citep{WuG2021}  &ResK    &Student ensemble         &DenseNet-40-12 (93.19)  &94.13   (0.94 $\uparrow$)\\\hline
PCL \citep{WuG2021}  &ResK    &Student ensemble         &VGG16 (93.96)            &94.74   (0.78 $\uparrow$)\\\hline
ACNs \citep{ZhangH2021}  &ResK    &ResNet14 (90.66)     &ResNet14 (90.66)         &92.09   (1.43 $\uparrow$)\\\hline
ACNs \citep{ZhangH2021}  &ResK    &VGG11 (91.25)     &VGG11 (91.25)               &92.65   (1.40 $\uparrow$)\\\hline
ACNs \citep{ZhangH2021}  &ResK    &AlexNet (73.24)     &AlexNet (73.24           &78.57   (5.33 $\uparrow$)\\\hline
\multicolumn{5}{|c|}{\textbf{Self-Distillation}} \\\hline
Methods             & Knowledge  & Teacher (baseline)         &Student (baseline)        &Accuracies  \\\hline
\cite{XuLiu2019}   &ResK, FeaK   &---              &ResNet32 (92.78)        &93.68        (0.90$\uparrow$)\\\hline
\cite{XuLiu2019}   &ResK, FeaK   &---              &DenseNe40(94.53)        &94.80       (0.27$\uparrow$)\\\hline
 \end{tabular}
\end{table*}

\begin{table*}[!ht]
  \centering
  \caption{Performance comparison of different knowledge distillation methods on CIFAR100. Note that  $\uparrow$ indicates the performance improvement of the student network learned by each method comparing with the corresponding baseline model.}\label{tbp100}
\begin{tabular}{|c|c|c|c|c|}
 \hline
\multicolumn{5}{|c|}{\textbf{Offline Distillation}} \\\hline
Methods               & Knowledge  & Teacher (baseline)  &Student (baseline)       &Accuracies  \\\hline
FSP \citep{Yim2017}   &RelK        &ResNet32 (64.06)     &ResNet14 (58.65)         &63.33        (4.68 $\uparrow$) \\\hline
FT \citep{KimPark2018} &FeaK       &ResNet110 (73.09)    &ResNet56 (71.96)         &74.48     (2.52 $\uparrow$)\\\hline
RKD \citep{ParkK2019} &RelK, FeaK  &ResNet50 (77.76)     &VGG11 (71.26)            &74.66    (3.40 $\uparrow$)\\\hline
IRG \citep{LiuC2019}  &RelK        &ResNet20 (78.40)     &ResNet20-x0.5 (72.51)    &74.64        (2.13 $\uparrow$)\\\hline
CCKD \citep{Peng2019}  &RelK, ResK &ResNet110 (---)      &ResNet20 (68.40)        &72.40        (4.00 $\uparrow$)\\\hline
KR \citep{LiuWen2019}  &FeaK       &ResNet32 (64.06)     &ResNet14 (58,65)         &63.95        (5.30 $\uparrow$)\\\hline
LKD \citep{LiX2020}    &RelK       &ResNet110 (75.76)     &ResNet20 (69.47)         &72.63        (3.16 $\uparrow$)\\\hline
LKD \citep{LiX2020}    &RelK      &WRN-40-2 (75.61)       &WRN-16-2 (73.10)         &75.44        (2.34 $\uparrow$)\\\hline
SSKD \citep{XuG2020}   &RelK, ResK &VGG13 (75.38)        &MobileNetV2 (65.79)      &71.53        (5.74 $\uparrow$)\\\hline
SSKD \citep{XuG2020}   &RelK, ResK &ResNet50 (79.10)    &MobileNetV2 (65.79)       &72.57        (6.78 $\uparrow$)\\\hline
FN \citep{XuK2020}     &FeaK      &ResNet110 (82.01)    &ResNet56 (81.73)          &82.23        (0.50 $\uparrow$)\\\hline
AdaIN \citep{YangECCV2020}          &FeaK         &WRN-40-4 (78.31)     &WRN-16-4 (75.68)           &78.25       (2.57 $\uparrow$)\\\hline
AdaIN \citep{YangECCV2020}          &FeaK         &ResNet34 (77.26)     &MobileNetV2 (68.36)        &70.66       (2.30 $\uparrow$)\\\hline
PAD-$L_{2}$ \citep{ZhangYECCV2020}  &FeaK         &ResNet18 (75.86)     &MobileNetV2 (68.16)        &74.06       (5.90 $\uparrow$)\\\hline
MGD \citep{YueECCV2020}             &FeaK         &WRN-28-4 (78.91)     &WRN-28-2 (75.12)           &78.82       (3.70 $\uparrow$)\\\hline
AE-KD \citep{Du2020}                &FeaK         &ResNet56 (---)       &ResNet20 (69.06)           &70.55       (1.49 $\uparrow$)\\\hline
JointRD \citep{LiG2020}             &FeaK         &ResNet18 (77.92)     &plain-CNN 18 (77.44)       &78.24       (0.80 $\uparrow$)\\\hline
TOFD \citep{ZhangL2020}             &FeaK         &ResNet152 (---)      &ResNet50  (77.42)          &84.74       (7.32 $\uparrow$)\\\hline
TOFD \citep{ZhangL2020}             &FeaK         &ResNet152 (---)      &ShuffleNetV2  (72.38)      &76.68       (4.30 $\uparrow$)\\\hline
CTKD \citep{ZhaoH2020}              &RelK, FeaK   &ResNet110  (72.65)   &ResNet20   (68.33)         &70.75       (2.42 $\uparrow$)\\\hline
CTKD \citep{ZhaoH2020}              &RelK, FeaK   &WRN-40-2 (75.42)     &WRN-16-2  (72.27)          &74.70       (2.43 $\uparrow$)\\\hline
SemCKD \citep{ChenD2021}            &FeaK         &ResNet-32x4 (79.42)  &VGG13  (74.82)             &79.43       (4.61 $\uparrow$)\\\hline
SemCKD \citep{ChenD2021}            &FeaK         &WRN-40-2 (75.61)     &MobileNetV2  (65.43)       &69.61       (4.18 $\uparrow$)\\\hline
SemCKD \citep{ChenD2021}            &FeaK         &VGG13 (74.64)        &ShuffleNetV2  (72.60)      &76.39       (3.79 $\uparrow$)\\\hline
RKD \citep{GaoM2021}                &FeaK         &ResNet34 (73.05)     &ResNet18  (68.06)          &72.82       (4.76 $\uparrow$)\\\hline
\multicolumn{5}{|c|}{\textbf{Online Distillation}} \\\hline
Methods                          & Knowledge  & Teacher (baseline)     &Student (baseline)         &Accuracies     \\\hline
Rocket-KD \citep{ZhouFan2018}    & FeaK       &WRN-40-1 (---)          &WRN-16-1 (56.30)           &67.00          (10.07 $\uparrow$)\\\hline
DML \citep{ZhangY2018}           & ResK       &WRN-28-10 (78.69)       &MobileNet (73.65)          &80.28, 77.39   (3.74 $\uparrow$)\\\hline
DML \citep{ZhangY2018}           & ResK      &MobileNet (73.65)        &ResNet32 (68.99)           &76.13, 71.10   (8.11 $\uparrow$)\\\hline
ONE \citep{Lanx2018}             &ResK       &ResNet32+ONE             &ResNet32 (68.82)           &73.39          (4.57 $\uparrow$)\\\hline
ONE \citep{Lanx2018}             &ResK       &ResNet110+ONE            &ResNet110 (74.67)          &78.38          (3.71 $\uparrow$)\\\hline
DCM \citep{YaoA2020}             & ResK     &WRN-28-10 (81.28)         &ResNet110 (73.45)          &82.18, 77.01   (3.56 $\uparrow$)\\\hline
DCM \citep{YaoA2020}             & ResK     &WRN-28-10 (81.28)         &MobileNet (73.70)          &83.17, 78.57  (4.87 $\uparrow$)\\\hline
KDCL \citep{Guo2020}      &ResK    &WRN-16-2 (72.20)    &ResNet32 (69.90)        &75.50, 74.30       (4.40 $\uparrow$)\\\hline
PCL \citep{WuG2021}       &ResK    &Student ensemble    &ResNet110 (76.21)       &79.98              (3.77 $\uparrow$)\\\hline
PCL \citep{WuG2021}       &ResK    &Student ensemble    &DenseNet-40-12 (71.03)  &73.09              (2.06 $\uparrow$)\\\hline
ACNs \citep{ZhangH2021}   &ResK    &ResNet14 (66.88)    &ResNet14 (66.88)        &68.40              (1.52 $\uparrow$)\\\hline
ACNs \citep{ZhangH2021}   &ResK    &VGG11 (67.38)       &VGG11 (67.38)           &70.11              (2.73 $\uparrow$)\\\hline
ACNs \citep{ZhangH2021}   &ResK    &AlexNet (39.45)     &AlexNet (39.45         &46.27              (6.82 $\uparrow$)\\\hline
\multicolumn{5}{|c|}{\textbf{Self-Distillation}} \\\hline
Methods                   & Knowledge  & Teacher (baseline)         &Student (baseline)        &Accuracies  \\\hline
\cite{XuLiu2019}          &ResK, FeaK  &---                         &DenseNet (74.80)          &76.32        (1.52$\uparrow$) \\\hline
SD~\citep{YangCXie2019}   &ResK        &---                         &ResNet32 (68.39)          &71.29     (2.90$\uparrow$) \\\hline
Tf-KD \citep{YuanL2019}   & ResK       &---                         &ResNet18 (75.87)          &77.10        (1.23$\uparrow$) \\\hline
Tf-KD \citep{YuanL2019}   & ResK       &---                         &ShuffleNetV2 (70.34)      &72.23        (1.89$\uparrow$) \\\hline
Tf-KD \citep{YuanL2019}   & ResK       &---                         &ResNeXt29 (81.03)         &82.08        (1.05$\uparrow$) \\\hline
CS-KD~\citep{Sukmin2019}  &ResK        &---                         &ResNet18 (75.29)          &78.01       (2.72$\uparrow$) \\\hline
 \end{tabular}
\end{table*}

\subsection{NAS-Based Distillation}
Neural architecture search (NAS), which is one of the most popular auto machine learning (or AutoML) techniques, aims to automatically identify deep neural models and adaptively learn appropriate deep neural structures. In knowledge distillation, the success of knowledge transfer depends on not only the knowledge from the teacher but also the architecture of the student. However, there might be a capacity gap between the large teacher model and the small student model, making it difficult for the student to learn well from the teacher. To address this issue, neural architecture search has been adopted to find the appropriate student architecture in  oracle-based \citep{Kang2020} and architecture-aware knowledge distillation \citep{LiuJia2019}. Furthermore, knowledge distillation is employed to improve the efficiency of neural architecture search, such as AdaNAS \citep{Macko2019}, NAS with distilled architecture knowledge \citep{LiP2020}, teacher guided search for architectures or TGSA~\citep{Bashivan2019}, and one-shot NAS ~\citep{PengH2020}. In TGSA, each architecture search step is guided to mimic the intermediate feature representations of the teacher network. The possible structures for the student are efficiently searched and the feature transfer is effectively supervised by the teacher.

\section{Performance Comparison}
\label{8}


Knowledge distillation is an excellent technique for model compression. Through capturing the teacher knowledge and using distillation strategies with teacher-student learning, it provides effective performance of the lightweight student model. Recently, many knowledge distillation methods focus on improving the performance, especially in image classification tasks. In this section, to clearly demonstrate the effectiveness of knowledge distillation, we summarize the classification performance of some typical KD methods on two popular image classification datasets.

The two datasets are CIFAR10 and CIFAR100 \citep{Krizhevsky2009} that are composed of $32\times32$ RGB images taken from 10 and 100 classes, respectively. Both have 50000 training images and 10000 testing images, and each class has the same numbers of training and testing images. For fair comparison, the experimental classification accuracy results (\%) of the KD methods are directly derived from the corresponding original papers, as shown in Table~\ref{tbp10} for CIFAR10 and Table~\ref{tbp100} for CIFAR100. We report the performance of different methods when using different types of knowledge, distillation schemes, and structures of teacher/student models. Specifically, the accuracies in parentheses are the classification results of the teacher and student models, which are trained individually. It should be noted that the pairs of accuracies of DML~\citep{ZhangY2018}, DCM~\citep{YaoA2020} and KDCL \citep{Guo2020} are the performance of teacher and student after online distillation.

From the performance comparison in Table~\ref{tbp10} and Table~\ref{tbp100}, several observations can be summarized as
\begin{enumerate}[$\bullet$]
\item Knowledge distillation can be realized on different deep models.
\item Model compression of different deep models can be achieved by knowledge distillation.
\item The online knowledge distillation through collaborative learning~\citep{ZhangY2018,YaoA2020} can significantly improve the performance of the deep models.
\item The self-knowledge distillation~\citep{YangCXie2019,YuanL2019,XuLiu2019,Sukmin2019} can well improve the performance of the deep models.
\item The offline and online distillation methods often transfer feature-based knowledge and response-based knowledge, respectively.
\item The performance of the lightweight deep models (student) can be improved by the knowledge transfer from the high-capacity teacher models.
\end{enumerate}
Through the performance comparison of different knowledge distillation methods, it can be easily concluded that knowledge distillation is an effective and efficient technique of compressing deep models.

\section{Applications}
\label{7}

As an effective technique for the compression and acceleration of deep neural networks, knowledge distillation has been widely used in different fields of artificial intelligence, including visual recognition, speech recognition, natural language processing (NLP), and recommendation systems. Furthermore, knowledge distillation also can be used for other purposes, such as the data privacy and as a defense against adversarial attacks.
This section briefly reviews applications of knowledge distillation.

\subsection{KD in Visual Recognition}
\label{7.1}

In last few years, a variety of knowledge distillation methods have been widely used for model compression in different visual recognition applications. Specifically, most of the knowledge distillation methods were previously developed for image classification \citep{LiZ2018,PengZ2019,Bagherinezhad2018,ChenW2018,WangJ2019,Mukherjee2019,ZhuM2019} and then extended to other visual recognition applications, including face recognition \citep{LuoP2016,Kongh2019,YanM2019,ShimingGe2019,WangM2018,WangM2019,Duong2019,WuHe2019,WangC2017,ZhangMECCV2020,WangXECCV2020}, image/video segmentation \citep{HeT2019,Mullapudi2019,DouQ2020,LiuY2019,SiamM2019,HouY2020,Bergmann2020}, action recognition \citep{LuoZ2018,HaoW2019,Thoker2019,Garcia2018,WangHu2018,WuM2019,ZhangS2020,CuiZ2020}, object detection \citep{LiQ2017,Shmelkov2017,CunECCV2020,WangT2019,HuangZ2020,WeiY2018,Hongyu2019,Chawla2021}, lane detection \citep{HouY2019}, person re-identification \citep{WuA2019}, pedestrian detection \citep{ShenJ2018}, facial landmark detection \citep{DongX2019}, pose estimation \citep{Nie2019,ZhangF2019,ZhaoLi2018}, video captioning \citep{PanB2020,ZhangZS2020}, person search \citep{Munjal2019,ZhangX2021}, image retrieval \citep{QingLiu2019}, shadow detection \citep{ChenZZhu2020}, saliency estimation \citep{LiJ2019}, depth estimation \citep{Pilzer2019,YeJ2019}, visual odometry \citep{Saputra2019},  text-to-image synthesis \citep{YuanM2020,TanH2021}, video classification \citep{ZhangPeng2018,Bhardwaj2019}, visual question answering \citep{MunJ2018,AdityaS2019} and anomaly detection \citep{Bergmann2020}. Since knowledge distillation in classification task is fundamental for other tasks, we briefly review knowledge distillation in challenging image classification settings, such as face recognition and action recognition.

Existing KD-based face recognition methods focus on not only efficient deployment but also competitive recognition accuracy \citep{LuoP2016,Kongh2019,YanM2019,ShimingGe2019,WangM2018,WangM2019,Duong2019,WangC2017,WangXECCV2020,ZhangMECCV2020}. Specifically, in \citep{LuoP2016}, the knowledge from the chosen informative neurons of top hint layer of the teacher network is transferred into the student network. A teacher weighting strategy with the loss of feature representations from hint layers was designed for knowledge transfer to avoid the incorrect supervision by the teacher \citep{WangM2018}. A recursive knowledge distillation method was designed by using a previous student network to initialize the next one \citep{YanM2019}. Since most face recognition methods perform the open-set recognition, i.e., the classes/identities on test set are unknown to the training set, the face recognition criteria are usually distance metrics between feature representations of positive and negtive samples, e.g., the angular loss in \citep{Duong2019} and the correlated embedding loss in \citep{WuHe2019}.




To improve low-resolution face recognition accuracy, the knowledge distillation framework is developed by using architectures between high-resolution face teacher and low-resolution face student for model acceleration and improved classification performance~\citep{ShimingGe2019,WangM2019,Kongh2019,Ge2020}. Specifically, \cite{ShimingGe2019} proposed a selective knowledge distillation method, in which the teacher network for high-resolution face recognition selectively transfers its informative facial features into the student network for low-resolution face recognition through sparse graph optimization. In \citep{Kongh2019}, cross-resolution face recognition was realized by designing a resolution invariant model unifying both face hallucination and heterogeneous recognition sub-nets. To get efficient and effective low resolution face recognition model, the multi-kernel maximum mean discrepancy between student and teacher networks was adopted as the feature loss \citep{WangM2019}. In addition, the KD-based face recognition can be extended to face alignment and verification by changing the losses in knowledge distillation \citep{WangC2017}.

Recently, knowledge distillation has been used successfully for solving the complex image classification problems
\citep{ZhuM2019,Bagherinezhad2018,PengZ2019,LiZ2018,ChenW2018,WangJ2019,Mukherjee2019}.
For incomplete, ambiguous and redundant image labels, the label refinery model through self-distillation and label progression is proposed to learn soft, informative, collective and dynamic labels for complex image classification \citep{Bagherinezhad2018}. To address catastrophic forgetting with CNN in a variety of image classification tasks, a learning without forgetting method for CNN, including both knowledge distillation and lifelong learning is proposed to recognize a new image task and to preserve the original tasks \citep{LiZ2018}. For improving image classification accuracy, \cite{ChenW2018} proposed the feature maps-based knowledge distillation method with GAN. It transfers knowledge from feature maps to a student.~Using knowledge distillation, a visual interpretation and diagnosis framework that unifies the teacher-student models for interpretation and a deep generative model for diagnosis is designed for image classifiers \citep{WangJ2019}. Similar to the KD-based low-resolution face recognition, \cite{ZhuM2019} proposed deep feature distillation for the low-resolution image classification, in which the output features of a student match that of teacher.

As argued in Section \ref{6.3}, knowledge distillation with the teacher-student structure can transfer and preserve the cross-modality knowledge. Efficient and effective action recognition under its cross-modal task scenarios can be successfully realized \citep{Thoker2019,LuoZ2018,Garcia2018,HaoW2019,WuM2019,ZhangS2020}. These methods are the examples of spatiotemporal modality distillation with a different knowledge transfer for action recognition. Examples include  mutual teacher-student networks \citep{Thoker2019}, multiple stream
networks \citep{Garcia2018}, spatiotemporal distilled dense-connectivity network \citep{HaoW2019}, graph distillation \citep{LuoZ2018} and multi-teacher to multi-student networks \citep{WuM2019,ZhangS2020}. Among these methods, the light-weight student can distill and share the knowledge information from multiple modalities stored in the teacher.

%

We summarize two main observations of distillation-based visual recognition applications, as follows.
\begin{enumerate}[$\bullet$]
\item Knowledge distillation provides efficient and effective teacher-student learning for a variety of different visual recognition tasks, because a lightweight student network can be easily trained under the guidance of the high-capacity teacher networks.
\item Knowledge distillation can make full use of the different types of knowledge in complex data sources, such as cross-modality data, multi-domain data and multi-task data and low-resolution data, because of flexible teacher-student architectures and knowledge transfer.
\end{enumerate}

\subsection{KD in NLP}
\label{7.2}

Conventional language models such as BERT are very time consuming and resource consuming with complex cumbersome structures. Knowledge distillation is extensively studied in the field of natural language processing (NLP), in order to obtain the lightweight, efficient and effective language models. More and more KD methods are proposed for solving the numerous NLP tasks \citep{LiuChen2019,Gordon2019,Haidar2019,YangZ2020,TangR2019,HuM2019,SunS2019,Nakashole2017,JiaoX2019,WangW2018,ZhouC2019,Sanh2019,Turc2019,Arora2019,Clark2019,KimY2016,MouL2016,LiuX2019,Hahn2019,TanX2019,Kuncoro2016,CuiJ2017,WeiH2019,Freitag2017,Shakeri2019,Aguilar2019,FuH2021,YangZ2020,ZhangS2021,ChenY2020,WangZ2021}. The existing NLP tasks using KD contain neural machine translation (NMT) \citep{Hahn2019,ZhouC2019,LiB2021,KimY2016,Gordon2019,TanX2019,WeiH2019,Freitag2017,ZhangS2021}, text generation \citep{ChenY2020,Haidar2019}, question answering system \citep{HuM2019,WangW2018,Arora2019,YangZ2020}, event detection \citep{LiuChen2019}, document retrieval \citep{Shakeri2019}, text recognition \citep{WangZ2021} and so on. Among these KD-based NLP methods, most of them belong to natural language understanding (NLU), and many of these KD methods for NLU  are designed as the task-specific distillation \citep{TangR2019,Turc2019,MouL2016} and multi-task distillation \citep{LiuX2019,YangZ2020,Sanh2019,Clark2019}. In what follows, we describe KD research works for neural machine translation and then for extending a typical multilingual representation model entitled bidirectional encoder representations from transformers ( or BERT) \citep{Devlin2019} in NLU.

In natural language processing, neural machine translation is the hottest application. {However, the existing NMT models with competitive performance is very large. To obtain lightweight NMT, there are many extended knowledge distillation methods for neural machine translation \citep{Hahn2019,ZhouC2019,KimY2016,Gordon2019,WeiH2019,Freitag2017,TanX2019}.
Recently, \cite{ZhouC2019} empirically proved the better performance of the KD-based non-autoregressive machine translation (NAT) model largely relies on its capacity and the distilled data via knowledge transfer. \cite{Gordon2019} explained the good performance of sequence-level knowledge distillation from the perspective of data augmentation and regularization.
In \citep{KimY2016}, the effective word-level knowledge distillation is extended to the sequence-level one in the sequence generation scenario of NMT. The sequence generation student model mimics the sequence distribution of the teacher. To overcome the multilingual diversity, \cite{TanX2019} proposed multi-teacher distillation, in which multiple individual models for handling bilingual pairs are teacher and a multilingual model is student.
To improve the translation quality, an ensemble of mutiple NMT models as teacher supervise the student model with a data filtering method \cite{Freitag2017}. To improve the performance of machine translation and machine reading tasks, \citep{WeiH2019} proposed a novel online knowledge distillation method, which addresses the unstableness of the training process and the decreasing performance on each validation set. In this online KD, the best evaluated model during training is chosen as teacher and updated by any subsequent better model. If the next model had the poor performance, the current teacher model would guide it.

As a multilingual representation model, BERT has attracted attention in natural language understanding  \citep{Devlin2019}, but it is also a cumbersome deep model that is not easy to be deployed. To address this problem, several lightweight variations of BERT (called BERT model compression) using knowledge distillation are proposed \citep{SunS2019,JiaoX2019,TangR2019,Sanh2019,WangW2020,LiuW2020,FuH2021}. \cite{SunS2019} proposed patient knowledge distillation for BERT model compression (BERT-PKD), which is used for sentiment classification, paraphrase similarity matching, natural language inference, and machine reading comprehension. In the patient KD method, the feature representations of the [CLS] token from the hint layers of teacher are transferred to the student. To accelerate language inference, \cite{JiaoX2019} proposed TinyBERT that is two-stage transformer knowledge distillation. It contains general-domain and task-specific knowledge distillation. For sentence classification and matching, \cite{TangR2019} proposed task-specific knowledge distillation from the BERT teacher model into a bidirectional long short-term memory network (BiLSTM). In \citep{Sanh2019}, a lightweight student model called DistilBERT with the same generic structure as BERT is designed and learned on a variety of tasks of NLP. In \citep{Aguilar2019}, a simplified student BERT is proposed by using the internal representations of a large teacher BERT via internal distillation.

Furthermore, some typical KD methods for NLP with different perspectives are represented below. For question answering, to improve the efficiency and robustness of machine reading comprehension,  \cite{HuM2019} proposed an attention-guided answer distillation method, which fuses generic distillation and answer distillation to avoid confusing answers. For a task-specific distillation \citep{Turc2019}, the performance of knowledge distillation with the interactions among pre-training, distillation and fine-tuning for the compact student model is studied. The proposed pre-trained distillation performs well in sentiment classification, natural language inference, textual entailment. For a multi-task distillation in the context of natural language understanding, \cite{Clark2019} proposed the single-multi born-again distillation, which is based on born-again neural networks \citep{Furlanello2018}. Single-task teachers teach a multi-task student. For multilingual representations, knowledge distillation transfers knowledge among the multi-lingual word embeddings for bilingual dictionary induction \citep{Nakashole2017}. For low-resource languages, knowledge transfer is effective across ensembles of multilingual models \citep{CuiJ2017}.

Several observations about knowledge distillation for natural language processing are summarized as follows.
\begin{enumerate}[$\bullet$]
\item Knowledge distillation provides efficient and effective lightweight language deep models. The large-capacity teacher model can transfer the rich knowledge from a large number of different kinds of language data to train a small student model, so that the student can quickly complete many language tasks with effective performance.
\item The teacher-student knowledge transfer can easily and effectively solve many multilingual tasks, considering that knowledge from multilingual models can be transferred and shared by each other.
\item In deep language models, the sequence knowledge can be effectively transferred from large networks into small networks.
\end{enumerate}

\subsection{KD in Speech Recognition}
\label{7.3}

In the field of speech recognition, deep neural acoustic models have attracted attention and interest due to their powerful performance. However, more and more real-time speech recognition systems are deployed in embedded platforms with limited computational resources and fast response time. The state-of-the-art deep complex models cannot satisfy the requirement of such speech recognition scenarios. To satisfy these requirements, knowledge distillation is widely studied and applied in many speech recognition tasks. There are many knowledge distillation systems for designing lightweight deep acoustic models for speech recognition~\citep{Chebotar2015,WongJ2016,ChanW2015,Price2016,Fukuda2017,BaiY2019,NgR2018,Albanie2018,LuL2017,ShiB2019a,Roheda2018,ShiB2019b,GaoL2019,Ghorbani2018,Takashima2018,Watanabe2017,ShiY2019,Asami2017,HuangM2019,ShenP2018,Perez2019,ShenP2019,Oord2018,Kwon2020,ShenP2020}.
In particular, these KD-based speech recognition applications have spoken language identification \citep{ShenP2018,ShenP2019,ShenP2020}, audio classification \citep{GaoL2019,Perez2019}, text-independent speaker recognition~\citep{NgR2018}, speech enhancement~\citep{Watanabe2017}, acoustic event detection \citep{Price2016, ShiB2019a,ShiB2019b}, speech synthesis \citep{Oord2018} and so on.

Most existing knowledge distillation methods for speech recognition, use teacher-student architectures to improve the efficiency and recognition accuracy of acoustic models \citep{ChanW2015,Watanabe2017,Chebotar2015,ShenP2019,LuL2017,ShenP2018,ShenP2020,GaoL2019,ShiY2019,ShiB2019a,Perez2019}. Using a recurrent neural network (RNN) for holding the temporal information from speech sequences, the knowledge from the teacher RNN acoustic model is transferred into a small student DNN model \citep{ChanW2015}. Better speech recognition accuracy is obtained by combining multiple acoustic modes. The ensembles of different RNNs with different individual training criteria are designed to train a student model through knowledge transfer~\citep{Chebotar2015}. The learned student model performs well on 2,000-hour large vocabulary continuous speech recognition (LVCSR) tasks in 5 languages. To strengthen the generalization of the spoken language identification (LID) model on short utterances, the knowledge of feature representations of the long utterance-based teacher network is transferred into the short utterance-based student network that can discriminate short utterances and perform well on the short duration utterance-based LID tasks \citep{ShenP2018}. To further improve the performance of short utterance-based LID, an interactive teacher-student online distillation learning is proposed to enhance the performance of the feature representations of short utterances \citep{ShenP2019}. The LID performance on short utterances is also improved by distilling internal representation knowledge of teacher on longer utterances into the one of student on short utterances \citep{ShenP2020}.

Meanwhile, for audio classification, a multi-level feature distillation method is developed and an adversarial learning strategy is adopted to optimize the knowledge transfer \citep{GaoL2019}. To improve noise robust speech recognition, knowledge distillation is employed as the tool of speech enhancement \citep{Watanabe2017}. In \citep{Perez2019}, a audio-visual multi-modal knowledge distillation method is proposed. knowledge is transferred from the teacher models on visual and acoustic data into a student model on audio data. In essence, this distillation shares the cross-modal knowledge among the teachers and students \citep{Perez2019,Albanie2018,Roheda2018}. For efficient acoustic event detection, a quantized distillation method is proposed by using both knowledge distillation and quantization \citep{ShiB2019a}. The quantized distillation transfers knowledge from a large CNN teacher model with better detection accuracy into a quantized RNN student model.

Unlike most existing traditional frame-level KD methods, sequence-level KD can perform better in some sequence models for speech recognition, such as connectionist temporal classification (CTC) \citep{WongJ2016,Takashima2018,HuangM2019}. In \citep{HuangM2019}, sequence-level KD is introduced into connectionist temporal classification, in order to match an output label sequence used in the training of teacher model and the input speech frames used in distillation. In \citep{WongJ2016}, the effect of speech recognition performance on frame-level and sequence-level student-teacher training is studied and a new sequence-level student-teacher training method is proposed. The teacher ensemble is constructed by using sequence-level combination instead of frame-level combination. To improve the performance of unidirectional RNN-based CTC for real-time speech recognition, the knowledge of a bidirectional LSTM-based CTC teacher model is transferred into a unidirectional LSTM-based CTC student model via frame-level KD and sequence-level KD \citep{Takashima2018}.

Moreover, knowledge distillation can be used to solve some special issues in speech recognition \citep{BaiY2019,Asami2017,Ghorbani2018}.  To overcome overfitting issue of DNN acoustic models when data are scarce, knowledge distillation is employed as a regularization way to train adapted model with the supervision of the source model \citep{Asami2017}. The final adapted model achieves better performance on three real acoustic domains. To overcome the degradation of the performance of non-native speech recognition, an advanced multi-accent student model is trained by distilling knowledge from the multiple accent-specific RNN-CTC models \citep{Ghorbani2018}. In essence, knowledge distillation in \citep{Asami2017,Ghorbani2018} realizes the cross-domain knowledge transfer. To solve the complexity of fusing the external language model (LM) into sequence-to-sequence model (Seq2seq) for speech recognition, knowledge distillation is employed as an effective tool to integrate a LM (teacher) into Seq2seq model (student) \citep{BaiY2019}. The trained Seq2seq model can reduce character error rates in sequence-to-sequence speech recognition.

In summary, several observations on knowledge distillation-based speech recognition can be concluded as follows.
\begin{enumerate}[$\bullet$]
\item The lightweight student model can satisfy the practical requirements of speech recognition, such as real-time responses, use of limited resources and high recognition accuracy.
\item Many teacher-student architectures are built on RNN models because of the temporal property of speech sequences. In general, the RNN models are chosen as the teacher, which can well preserve and transfer the temporal knowledge from real acoustic data to a student model.
\item Sequence-level knowledge distillation can be well applied to sequence models with good performance. In fact, the frame-level KD always uses the response-based knowledge, but sequence-level KD usually transfers the feature-based knowledge from hint layers of teacher models.
\item Knowledge distillation using teacher-student knowledge transfer can easily solve the cross-domain or cross-modal speech recognition in applications such as multi-accent and multilingual speech recognition.
\end{enumerate}

\subsection{KD in Other Applications}
\label{7.4}

The full and correct leverages of external knowledge, such as in a user review or in images, play a very important role in the effectiveness of deep recommendation models. Reducing the complexity and improving the efficiency of deep recommendation models is also very necessary. Recently, knowledge distillation has been successfully applied in recommender systems for deep model compression and acceleration \citep{ChenZhang2018,TangJ2018,PanY2019}. In \citep{TangJ2018}, knowledge distillation is first introduced into the recommender systems and called ranking distillation because the recommendation is expressed as a ranking problem. \cite{ChenZhang2018} proposed an adversarial knowledge distillation method for efficient recommendation. A teacher as the right review predication network supervises the student as user-item prediction network (generator). The student learning is adjusted by adversarial adaption between teacher and student networks. Unlike distillation in \citep{ChenZhang2018,TangJ2018}, \cite{PanY2019} designed a enhanced collaborative denoising autoencoder (ECAE) model for recommender systems via knowledge distillation to capture useful knowledge from user feedbacks and to reduce noise. The unified ECAE framework contains a generation network, a retraining network and a distillation layer that transfers knowledge and reduces noise from the generation network.

Using the natural characteristic of knowledge distillation with teacher-student architectures, knowledge distillation is used as an effective strategy to solve adversarial attacks or perturbations of deep models \citep{Papernot2018,Ross2018,Micah2019,Gil2019} and the issue of the unavailable data due to the privacy, confidentiality and security concerns \citep{Raphael2017,Papernot2017,WangBao2019,BaiH2019,Vongku2019}. To be specific, the perturbations of the adversarial samples can be overcome by the robust outputs of the teacher networks via distillation \citep{Ross2018,Papernot2018}. To avoid exposing the private data, multiple teachers access subsets of the sensitive or unlabelled data and supervise the student \citep{Papernot2017,Vongku2019}. To address the issue of privacy and security, the data to train the student network is generated by using the layer activations or layer spectral activations of the teacher network via data-free distillation \citep{Raphael2017}. To protect data privacy and prevent intellectual piracy, \cite{WangBao2019} proposed a private model compression framework via knowledge distillation. The student model is applied to public data while the teacher model is applied to both sensitive and public data. This private knowledge distillation adopts privacy loss and batch loss to further improve privacy. To consider the compromise between privacy and performance, \cite{BaiH2019} developed a few shot network compression method via a novel layer-wise knowledge distillation with few samples per class. Of course, there are other special interesting applications of knowledge distillation, such as neural architecture search \citep{Macko2019,Bashivan2019}, interpretability of deep neural networks \citep{LiuX2018}, and federated learning~\citep{Bistritz2020,LinT2020,SeoH2020,HeC2020}.

\section{Conclusion and Discussion}
\label{9}

Knowledge distillation and its applications have aroused considerable attention in recent few years. In this paper, we present a comprehensive review on knowledge distillation, from the perspectives of knowledge, distillation schemes, teacher-student architectures, distillation algorithms, performance comparison and applications. Below, we discuss the challenges of knowledge distillation and provide some insights on the future research of knowledge distillation.

\subsection{Challenges}
\label{8.1}

For knowledge distillation, the key is to 1) extract rich knowledge from the teacher and 2) to transfer the knowledge from the teacher to guide the training of the student. Therefore, we discuss the challenges in knowledge distillation from the followings aspects: the quality of knowledge, the types of distillation, the design of the teacher-student architectures, and the theory behind knowledge distillation.

Most KD methods leverage a combination of different kinds of knowledge, including response-based, feature-based, and relation-based knowledge. Therefore, it is important to know the influence of each individual type of knowledge and to know how different kinds of knowledge help each other in a complementary manner. For example, the response-based knowledge has a similar motivation to label smoothing and the model regularization \citep{Kims2017,Muller2019,Ding2019}; The featured-based knowledge is often used to mimic the intermediate process of the teacher and the relation-based knowledge is used to capture the relationships across different samples.  To this end, it is still challenging to model different types of knowledge in a unified and complementary framework. For example, the knowledge from different hint layers may have different influences on the training of the student model: 1) response-based knowledge is from the last layer; 2) feature-based knowledge from the deeper hint/guided layers may suffer from over-regularization~\citep{Romero2015}.}

How to transfer the rich knowledge from the teacher to a student is a key step in knowledge distillation. Generally, the existing distillation methods can be categorized into offline distillation, online distillation and self distillation. Offline distillation is usually used to transfer knowledge from a complex teacher model, while the teacher model and the student model are comparable in the settings of online distillation and self distillation. To improve the efficacy of knowledge transfer, the relationships between the model complexity and existing distillation schemes or other novel distillation schemes~\citep{SunL2021} should be further investigated.

Currently, most KD methods focus on new types of knowledge or distillation loss functions, leaving the design of the teacher-student architectures poorly investigated~\citep{Nowak2018,Crowley2018,Kang2020,LiuJia2019,Ashok2018,LiuPeng2019}. In fact, apart from the knowledge and distillation algorithms, the relationship between the structures of the teacher and the student also significantly influences the performance of knowledge distillation. For example, on one hand, some recent works find that the student model can learn little from some teacher models due to the model capacity gap between the teacher model and the student model~\citep{ZhangL2019,Kang2020}; On the other hand, from some early theoretical analysis on the capacity of neural networks, shallow networks are capable of learning the same representation as deep neural networks~\citep{Ba2014}. Therefore, the design of an effective student model or construction of a proper teacher model are still challenging problems in knowledge distillation.

Despite a huge number of the knowledge distillation methods and applications, the understanding of knowledge distillation including theoretical explanations and empirical evaluations remains insufficient \citep{Lopez2016,Phuong2019,ChoJ2019}. For example, distillation can be viewed as a form of learning with privileged information~\citep{Lopez2016}.
The assumption of linear teacher and student models enables the study of the theoretical explanations of characteristics of the student learning via distillation~\citep{Phuong2019}. Furthermore, some empirical evaluations and analysis on the efficacy of knowledge distillation were performed by \cite{ChoJ2019}. However, a deep understanding of generalizability of knowledge distillation, especially how to measure the quality of knowledge or the quality of the teacher-student architecture, is still very difficult to attain.

\subsection{Future Directions}
\label{8.2}

In order to improve the performance of knowledge distillation, the most important factors include what kind of teacher-student network architecture, what kind of knowledge is learned from the teacher network, and where is distilled into the student network.

The model compression and acceleration methods for deep neural networks usually fall into four different categories, namely parameter pruning and sharing, low-rank factorization, transferred compact convolutional filters and knowledge distillation \citep{ChengY2018}. In existing knowledge distillation methods, there are only a few related works discussing the combination of knowledge distillation and other kinds of compressing methods. For example, quantized knowledge distillation, which can be seen as a parameter pruning method, integrates network quantization into the teacher-student architectures \citep{Polino2018,Mishra2018,WeiY2018}. Therefore, to learn efficient and effective lightweight deep models for the deployment on portable platforms, the hybrid compression methods via both knowledge distillation and other compressing techniques are necessary, since most compressing techniques require a re-training/fine-tuning process. Furthermore, how to decide the proper orders for applying different compressing methods will be an interesting topic for future study.

Apart from model compression for acceleration for deep neural networks, knowledge distillation also can be used in other problems because of the natural characteristics of knowledge transfer on the teacher-student architecture. Recently, knowledge distillation has been applied to  the data privacy and security \citep{WangBao2019}, adversarial attacks of deep models \citep{Papernot2018}, cross-modalities \citep{Gupta2016}, multiple domains \citep{Asami2017}, catastrophic forgetting \citep{LeeKK2019}, accelerating learning of deep models \citep{ChenT2016}, efficiency of neural architecture search \citep{Bashivan2019}, self-supervision \citep{Noroozi2018}, and data augmentation \citep{LeeHwang2019,Gordon2019}. Another interesting example is that the knowledge transfer from the small teacher networks to a large student network can accelerate the student learning \citep{ChenT2016}. This is very quite different from vanilla knowledge distillation. The feature representations learned from unlabelled data by a large model  can also supervise the target model via distillation \citep{Noroozi2018}. To this end, the extensions of knowledge distillation for other purposes and applications might be a meaningful future direction.

The learning of knowledge distillation is similar to the human beings learning. It can be practicable to popularize the knowledge transfer to the classic and traditional machine learning methods \citep{ZengZhou2019,Gong2019,Yous2018,Gong2017}. For example, traditional two-stage classification is felicitous cast to a single teacher single student problem based on the idea of knowledge distillation \citep{ZengZhou2019}. Furthermore, knowledge distillation can be flexibly deployed to various excellent learning schemes, such as the adversarial learning \citep{LiuFusi2018}, auto machine learning \citep{Macko2019,Fakoor2020}, label noise filtering learning \citep{XiaS2018}, lifelong learning \citep{Zhai2019}, and reinforcement learning \citep{Ashok2018,XuZ2020,ZhaoC2020}. Therefore, it will be useful to integrate knowledge distillation with other learning schemes for practical challenges in the future.

\end{document}